\documentclass[11pt]{article}

\usepackage[english]{babel}
\usepackage[T1]{fontenc}

\usepackage[a4paper,top=3cm,bottom=2cm,left=3cm,right=3cm,marginparwidth=1.75cm]{geometry}

\usepackage{amsmath}
\usepackage{graphicx}
\usepackage[colorlinks=true, allcolors=blue, pdfencoding=auto]{hyperref}

\usepackage{graphicx} % more modern
\usepackage[noend]{algorithm}
\usepackage{algorithmic}

\usepackage{mathrsfs}
\usepackage{amsmath}
\usepackage{amssymb}

\usepackage[utf8]{inputenc}
\usepackage{amsthm}
\usepackage{thmtools, thm-restate}

\declaretheorem{theorem}
\declaretheorem{lemma}
\declaretheorem{corollary}

\newtheorem*{theorem*}{Theorem}

\usepackage{lipsum}% http://ctan.org/pkg/lipsum

\usepackage{subcaption}% http://ctan.org/pkg/subcaption
\DeclareCaptionSubType*{algorithm}

\DeclareCaptionLabelFormat{alglabel}{Table~#2}
\renewcommand{\algorithmiccomment}[1]{\hspace{2em}// #1} 
\floatname{algorithm}{Procedure}
\renewcommand{\algorithmicrequire}{\textbf{Input:}}
\renewcommand{\algorithmicensure}{\textbf{Output:}}

\usepackage{enumitem}
\usepackage{ragged2e}

\usepackage{changepage}

\usepackage{booktabs, adjustbox}
\usepackage{multirow}
\usepackage{threeparttable}
\usepackage{graphicx}
\usepackage{array}
\usepackage{tabularx}

\graphicspath{ {figs/} }
\usepackage{caption}
\usepackage{wrapfig,lipsum,booktabs}

\usepackage[toc,page]{appendix}

\usepackage{placeins}

\usepackage{cleveref}

\renewcommand{\H}{\mathcal{K}}

\newcommand{\PH}{PH}

\newcommand{\PHmP}{P^{\frac{1}{2}}H_mP^{\frac{1}{2}}}

\newcommand{\bx}{{\pmb x}}
\newcommand{\by}{{\pmb y}}
\newcommand{\bz}{{\pmb z}}
\newcommand{\bb}{{\pmb b}}
\newcommand{\be}{{\pmb e}}
\newcommand{\bg}{{\pmb g}}

\newcommand{\bv}{{\pmb v}}

\newcommand{\balpha}{{\pmb \alpha}}
\newcommand{\bbeta}{{\pmb \beta}}
\newcommand{\bomega}{{\pmb \omega}}
\newcommand{\HH}{{\mathcal H}}

\let\oldphi\phi
\renewcommand\phi{\operatorname{\oldphi}}
\newcommand{\rmb}{\operatorname{b}}
\newcommand{\rmd}{\mathrm{d}}
\newcommand{\rme}{\operatorname{e}}

\newcommand{\rmk}{\operatorname{k}}
\newcommand{\rmH}{\mathcal{K}}
\newcommand{\rmI}{\operatorname{I}}
\newcommand{\rmP}{\operatorname{P}}

\newcommand{\rmS}{\operatorname{S}}
\newcommand{\rmV}{\operatorname{V}}
\newcommand{\rmZ}{\operatorname{Z}}

\newcommand\defeq{\mathrel{\overset{\makebox[0pt]{\mbox{\normalfont\tiny\sffamily def}}}{=}}}
\newcommand\norm[1]{\left\lVert#1\right\rVert}

\usepackage{mathtools}

\DeclarePairedDelimiter\floor{\lfloor}{\rfloor}

\usepackage{mathrsfs}
\DeclareMathAlphabet{\mathpzc}{OT1}{pzc}{m}{it}

\theoremstyle{definition}
\newtheorem{remark}{Remark}[section]

\usepackage{authblk}
\newcommand{\newtxtblock}{}

\usepackage{csquotes}

\usepackage{caption} 
\title{Diving into the shallows: a computational perspective on large-scale shallow learning}
\author{Siyuan Ma, Mikhail Belkin}
\affil{Department of Computer Science and Engineering}
\affil{The Ohio State University}
\affil{\textit{\{masi,mbelkin\}@cse.ohio-state.edu}}

\date{\today}
\begin{document}
\maketitle

\vspace{-10mm}
\begin{abstract}

Remarkable recent success of deep neural networks has not been easy to analyze theoretically. It has been  particularly hard to disentangle relative significance of architecture and optimization in achieving accurate classification on large datasets. On the other hand, shallow methods (such as kernel methods) have  encountered obstacles in scaling to large data, despite excellent performance on smaller datasets, and extensive theoretical analysis. Practical large-scale optimization methods, such as variants of gradient descent, used so successfully in deep learning, seem to produce below par results when applied to kernel methods. 
 
In this paper we first identify a basic limitation in gradient descent-based optimization methods when used in conjunctions with smooth kernels. An analysis based on the spectral properties of the kernel demonstrates that only a vanishingly  small portion of the function space is {\it reachable} after a polynomial number of gradient descent iterations. This lack of approximating power drastically limits  gradient descent for a fixed computational budget leading to serious over-regularization/underfitting. The issue is purely algorithmic, persisting even in the limit of infinite data.

To address this shortcoming in practice, we introduce EigenPro iteration, based on a simple and direct preconditioning scheme using a small number of approximately computed eigenvectors. 
It can also be viewed as learning a new kernel optimized for gradient descent. 
It turns out that injecting this small (computationally inexpensive and SGD-compatible) amount of approximate second-order information leads to major improvements in convergence. For large data, this translates into significant performance  boost over the standard   kernel methods. In particular, we are able to consistently match or improve the state-of-the-art results recently reported in the literature with a small fraction of their computational budget. 

Finally, we feel that these results show a need for a broader computational perspective on modern large-scale learning to complement more traditional statistical and convergence analyses.   
In particular,  many phenomena of large-scale high-dimensional inference are best understood in terms of optimization on infinite dimensional Hilbert spaces, where standard algorithms can sometimes have properties at odds with finite-dimensional intuition.   A systematic analysis concentrating on the approximation power of such algorithms within a budget of computation may lead to progress both in theory and practice.

\end{abstract}

% \section{to do}

% % 1.kernel methods
% % 2. $\eta$
% % 3. theoretical analysis for fixed $P$. 
% 4. Rewrite motivation.
% \input{fitting.tex}

% \cite{cheng2011arccosine, huang2014kernel, dai2014scalable, lu2014scale} have showed that kernel methods are able to match and even beat the performance of DNN on small datasets. 

\section{Introduction}

In recent years we have witnessed remarkable advances in many areas of artificial intelligence.  %from computer vision to speech recognition to machine translation. 
In large part this progress has been due to the success of machine learning methods, notably deep neural networks, applied to very large datasets. These networks are typically trained using variants of stochastic gradient descent (SGD), allowing training on large data using modern hardware. Despite intense recent research and significant progress toward understanding SGD and deep architectures, it has not been easy to understand the underlying causes of that success. Broadly speaking, it can be attributed to (a) the structure of the function space represented by the network or (b) the properties of the optimization algorithms used. While these two aspects of learning are intertwined, they are distinct and there is hope that they may be disentangled.
%In particular, we note some intriguing experimental evidence on the (unreasonable?) effectiveness of  stochastic gradient descent in fitting the data with neural nets~\cite{zhang2016understanding}). 

As learning in deep neural networks is still largely resistant to theoretical analysis, progress both in theory and practice can be made by exploring the limits of shallow methods on large datasets.  Shallow methods, such as kernel methods, are a subject of an extensive and diverse literature. Theoretically, kernel machines  are known to be universal learners, capable of learning nearly arbitrary functions given a sufficient number of examples~\cite{shawe2004kernel,steinwart2008support}. 
Kernel methods are easily implementable and show state-of-the-art performance on smaller datasets (see~\cite{cheng2011arccosine, huang2014kernel, dai2014scalable, lu2014scale, may2017kernel} for some comparisons with DNN's). On the other hand,  
there has been significantly less progress in applying  these methods to large modern data\footnote{However, see~\cite{huang2014kernel, may2017kernel} for some notable successes.}.  
The goal of this work is to make a step toward understanding the subtle interplay between architecture and optimization for shallow algorithms and to take  practical steps to improve performance of kernel methods on large data.

The paper consists of two main parts. 
First, we identify a basic underlying limitation of using  gradient descent-based methods in conjunction with smooth  kernels typically used in machine learning. We  show that only very smooth  functions can be well-approximated after polynomially many steps of gradient descent. On the other hand, a less smooth target function cannot be approximated within $\epsilon$ using any polynomial number $P(1/\epsilon)$ steps of gradient descent for kernel regression. 
This phenomenon is a result of the fast spectral decay of  smooth kernels and  can be readily understood in terms of the spectral structure of the gradient descent operator in the least square regression/classification setting, which is the focus of our discussion.
Note the marked contrast with the standard analysis of gradient descent for convex optimizations problems, requiring at most $O(1/\epsilon)$ steps to get an $\epsilon$-approximation of a minimum. The difference is  due to the infinite (or, in practice,  very high) dimensionality of the target space and the fact that the minimizer of a convex functional is not generally an element of the same space. 

A direct consequence of this theoretical analysis is
slow convergence of gradient descent methods for high-dimensional regression, resulting in severe  over-regularization/underfitting and suboptimal performance for less smooth functions. These functions are arguably very common in practice, at least in the classification setting, where we expect sharp transitions or even discontinuities near the class boundaries. We give some examples on real data showing that the number of steps of gradient descent needed to obtain near-optimal classification is indeed very large even for smaller problems. 
% results from the fast spectral decay of  smooth kernels typically used in machine learning. It can be readily understood in terms of the spectral structure of the gradient descent operator in the least square regression/classification setting, which is the focus of our discussion.    
This shortcoming of gradient descent is purely  algorithmic and is not related to the sample complexity  of the data, persisting even in the limit of infinite data. It is also not an  intrinsic flaw of  kernel architectures,
%which is capable of approximation arbitrary functions but with parameter setting potentially requiring very large (nearly exponential) number of gradient descent iterations.
which are capable of approximating arbitrary functions but  potentially require a very large  number of gradient descent steps.  
%While non-smooth kernels (e.g., the Laplacian kernel) are available they typically 
The issue is particularly serious for large data, where direct second order methods cannot be used due to the computational constraints. Indeed, even for a dataset with only $10^6$ data points, practical direct solvers  require cubic, on the order of $10^{18}$ operations, weeks of computational time for a fast processor/GPU. 
%On the other hand, gradient descent-type methods hold the promise of a small number of quadratic,  $O(10^{12})$, matrix-vector multiplications, a much more manageable task.
While many approximate second-order methods are available, they rely on low-rank approximations and, as we discuss below, also lead to over-regularization as important information is contained in eigenvectors with very small eigenvalues typically discarded in such approximations.

In the second part of the paper we address this problem by proposing EigenPro iteration (see 
\href{http://www.github.com/EigenPro}{github.com/EigenPro} for the code), a direct and simple method to alleviate slow convergence resulting from fast eigen-decay for kernel (and covariance) matrices. EigenPro is a preconditioning scheme based on approximately computing a small number of top eigenvectors to modify the spectrum of these matrices. It can also be viewed as constructing a new kernel, specifically optimized for gradient descent. While EigenPro uses approximate second-order information, it is only employed to modify first-order gradient descent leading to the same mathematical solution (without introducing a bias). Moreover, only one second-order problem at the start of the iteration  needs to be solved. 
EigenPro requires  only  a small overhead per iteration compared to standard gradient descent and  is also fully compatible with SGD. We analyze the step size in the SGD setting and provide a range of experimental results for different kernels and parameter settings showing consistent  acceleration by a factor from five to over thirty   over the standard methods, such as Pegasos~\cite{shalev2011pegasos} for a range of datasets and settings. 
For large datasets, when the computational budget is limited,  that acceleration translates into significantly improved accuracy and/or computational efficiency. It also obviates the need for complex computational resources such as supercomputer nodes or AWS clusters typically used with other methods.  
In particular, we are able to  improve or match  the state-of-the-art recent results for large datasets in the kernel literature at a small fraction of their reported  computational budget, using a single GPU.

% We proceed to show how the issue of {\it limited reach} of gradient descent can be alleviated by using  a simple preconditioning scheme, which we call EigenPro iteration,  based on approximate computation of top eigenvectors.  We demonstrate its  application in the stochastic gradient descent setting and provide a number of experiments on large datasets, matching or improving  on the existing state of the art for kernel methods. We note that most of those results in the literature  utilize large computational resources, such as supercomputer nodes, while we require  a modest computational budget of a few GPU-hours.
 
Finally, we  note that in the large data setting, we are limited to a small number of iterations of gradient descent and certain approximate second-order computations. 
Thus, investigations  of algorithms based on  the space of functions that can be approximated within a fixed computational budget of these operations (defined in terms of the input data size) reflect the realities of modern large-scale machine learning more accurately than the more traditional analyses of convergence rates. Moreover, many aspects of modern inference are best reflected by an infinite dimensional optimization problem whose properties are sometimes different from the standard finite-dimensional results. 
Developing careful analyses and insights into these issues will no doubt 
 result in significant  pay-offs both in theory and in practice.

% While  discussions along these lines exist in the literature (cf.~\cite{bousquet2008tradeoffs}), they are not yet common.  In this paper we see that carefully aggregating first and second-order  information can result in large pay-offs both in theory and in practice. 

% It is also interesting to compare  this notion with the  idea of algorithmic luckiness~\cite{herbrich2002algorithmic}, which aims to analyze the space  of outputs for a specific algorithm given a fixed {\it data} (as opposed to computational) budget. 

\section{Gradient descent for shallow methods}
\label{sec:preliminary}
{\bf Shallow methods.} In the context of this paper, shallow methods denote the family of algorithms consisting of a (linear or non-linear) {\it feature map} $\phi:\mathbb{R^N}\to \mathcal{H}$ to a (finite or infinite-dimensional)  Hilbert space $\mathcal{H}$ followed by a linear regression/classification algorithm. This is a simple yet powerful setting amenable to theoretical analysis. In particular, it includes the class of kernel methods, where  the feature map typically takes us from finite dimensional input to an infinite dimensional Reproducing Kernel Hilbert Space (RKHS). In what follows we will employ the square loss which significantly simplifies the analysis and leads to efficient and competitive algorithms. 

% While the setting is that of regression, in our experience and as reported in the literature, the square loss is competitive with other loss functions for high-dimensional classification problems.  
%In our experience without sacrificing classification accuracy on most real-world data, when used in conjunction with kernel methods. 
\newtxtblock
%Furthermore, to simplify the analysis, we will consider square loss.

\noindent{\bf Linear regression.} 
Consider  $n$ labeled data points $\{(\bx_1, y_1), ..., (\bx_n, y_n) \in \mathcal{H}\times \mathbb{R}\}$.
%\footnote{We adopt bold-face letters to denote vectors.}. 
To simplify the notation let us assume that the feature map has already been applied to the data, i.e., $\bx_i = \phi(\bz_i)$. Least square linear regression aims to recover  the parameter vector $\alpha^*$ that  minimize the empirical loss as follows:
\begin{equation} \label{eq:lls}
L(\balpha) \defeq \frac{1}{n} \sum_{i=1}^{n} %{(\frac{1}{\sqrt{d}} 
(\langle\balpha, \bx_i \rangle_\HH - y_i)^2
\end{equation}
\begin{equation} \label{eq:empirical-loss-optimal}
\balpha^* = \arg\min_{\balpha \in \HH} {L(\balpha)}
\end{equation}
%It looks for $\balpha \in \mathbb{R}^d$ that minimizes the empirical loss, defined as
When $\balpha^*$ is not uniquely defined, we can  choose the smallest norm solution. We do not include the typical regularization term, $\lambda\|\alpha\|_\HH^2$ for reasons which will become clear shortly\footnote{We will argue that explicit regularization is  rarely needed when using kernel methods for large data as available computational methods tend to over-regularize even without  additional regularization.}.

Minimizing the empirical loss is related to solving a linear system of equations. Define the data matrix\footnote{We will take some liberties with infinite dimensional objects by sometimes treating them as vectors/matrices and writing $\langle \balpha, \bx\rangle_\HH$ as $\balpha^T\bx$.} $X \defeq (\bx_1, ..., \bx_n)^T$ and the label vector $\by \defeq (y_1, ..., y_n)^T$, as well as the (non-centralized) covariance matrix/operator,
\begin{equation} \label{eq:linear-hessian}
H \defeq \frac{2}{ n} \sum_{i=1}^{n}{\bx_i \bx_i^T} =\frac{2}{ n} X^T X
\end{equation}
Rewrite the loss as $L(\balpha) = \frac{1}{n} \norm{X\balpha - \by}^2_2$. Since $\nabla L(\balpha)\mid_{\balpha = \balpha^*} = 0$, minimizing $L(\balpha)$ is equivalent to solving the  linear system 
\begin{equation} \label{eq:linear}
H\balpha - \bb = 0
\end{equation}
with  $\bb = X^T \by$.
When $d=\dim(\HH)<\infty$, the time complexity of solving the linear system in Eq.~\ref{eq:linear} directly (using  Gaussian elimination or other methods typically employed in practice) is $O(d^3)$.

%{\bf Remark.}
\begin{remark}
For kernel methods we frequently have $d=\infty$. Instead of solving Eq.~\ref{eq:linear}, one solves the dual $n\times n$ system 
$ K \balpha - \by=0 $
where $K \defeq [k(\bz_i, \bz_j)]_{i,j = 1,\ldots,n}$ is the kernel matrix corresponding to the kernel function $k(\cdot, \cdot)$.
The solution  can be written as $\sum_{i=1}^{n} k(\bz_i,\cdot) \balpha(\bz_i)$. A direct solution  requires $O(n^3)$ operations. %where $k$ is the underlying kernel function. %\newtxtblock
\end{remark}

%%%%
\noindent {\bf Gradient descent (GD).} While linear systems of equations can be solved by direct methods, their computational demands make them impractical for large data. 
% Even for a dataset with only $10^6$ data points, direct second order solvers  require on the order of $10^{18}$ operations, weeks of computational time for a fast GPU. 
On the other hand, gradient descent-type iterative methods hold the promise of a small number of  $O(n^2)$ matrix-vector multiplications, a much more manageable task. Moreover, these methods can typically be used in a stochastic setting, reducing computational requirements and allowing for very efficient GPU implementations. These schemes are adopted in popular kernel methods implementations such as NORMA~\cite{kivinen2004online}, SDCA~\cite{hsieh2008dual}, Pegasos~\cite{shalev2011pegasos},
and DSGD~\cite{dai2014scalable}.

For linear systems of equations gradient descent takes a particularly simple form known as Richardson iteration~\cite{richardson1911approximate}. It is given by 
\begin{equation}\label{eq:richardson}
\balpha^{(t+1)} = \balpha^{(t)} - \eta (H \balpha^{(t)}-\bb)
\end{equation}
We see that
$$
\balpha^{(t+1)} - \balpha^*= (\balpha^{(t)} - \balpha^*) -\eta H (\balpha^{(t)} -  \balpha^*)
$$
% $$ 
% \balpha^{(t+1)} - \balpha^* = (I - \eta H)(\balpha^{(t)} - \balpha^*)
% $$
and thus
\begin{equation}
\label{eq:operator_grad_descent}
\balpha^{(t+1)} - \balpha^* = (I - \eta H)^{t}(\balpha^{(1)} - \balpha^*)
\end{equation}
It is easy to see that for convergence of $\balpha^{t}$ to $\balpha^*$ as $t \to \infty$ we need to ensure\footnote{In general $\eta$ is chosen as a function of $t$. However, in the least squares setting $\eta$ can be chosen to be a constant as the Hessian matrix does not change.} that $\|I - \eta H\| \le 1 $. It follows that  $0<\eta < 2/\lambda_1(H)$.

\begin{remark}
When $H$ is finite dimensional the inequality has to be strict. In infinite dimension convergence is possible even if $\|I - \eta H\| = 1$ as long as each eigenvalue of   $I -\eta H$ is strictly smaller than one in absolute value. That will be the case for kernel integral operators.
\end{remark}

It is now easy to describe the {\it computational reach} of gradient descent $\mathcal{CR}_t$, i.e. the set of vectors which can be $\epsilon$-approximated  by gradient descent after $t$ steps 
$$
\mathcal{CR}_t(\epsilon) \defeq \{\bv \in \HH, s.t. \|(I - \eta H)^{t}\bv\|<\epsilon \|\bv\|\}
$$
It is important to note that any $\bb \notin \mathcal{CR}_t(\epsilon)$ cannot be $\epsilon$-approximated by gradient descent in  less than $t+1$ iterations.  
\begin{remark}
%{\bf Remark 1:} 
We typically care about the quality of the solution $\|H\balpha^{(t)} - \bb\|$, rather than the error estimating the parameter vector $\|\balpha^{(t)}- \balpha^*\|$ where $\balpha^*=H^{-1}\bb$. Thus (noticing that $H$ and $(I - \eta H)^{t}$ commute), we get $\|(I - \eta H)^{t}\bv\| = \|H^{-1}(I - \eta H)^{t}H\bv\|$ in the definition.
\end{remark}

\begin{remark} [Initialization]
%\noindent {\bf Remark 2:}
We assume that the initialization $\balpha^{(1)}=0$. Choosing a different starting point will not significantly change the analysis unless second order information is incorporated in the initialization conditions\footnote{This is different for non-convex methods where different initializations may result in convergence to different local minima.}. We also note that if $H$ is not full rank, gradient descent will converge to the minimum norm solution of Eq.~\ref{eq:linear}. 
\end{remark}

\begin{remark} [Infinite dimensionality]
%\noindent {\bf Remark 3:} 
Some care needs to be taken when $K$ is infinite-dimensional. In particular, the space of parameters $\alpha = K^{-1}\HH$ and $\HH$ are very different  spaces when $K$ is an integral operator. The space of parameters is in fact a space of distributions (generalized functions). Sometimes that can be addressed by using $K^{1/2}$ instead of $K$, as $K^{-1/2} \HH = L^2(\Omega)$. 
%As this distinction does not change the nature or add clarity to the discussion below,
%and given the limitations of the ICML format
%we will suppress the space notation. 
\end{remark}

To get a better idea of the space $\mathcal{CR}_t(\epsilon)$ 
consider the eigendecomposition of $H$. Let  $\lambda_1 \geq \lambda_2 \geq \ldots \ge 0$ be its eigenvalues and $\be_1,\be_2,\ldots$  the corresponding eigenvectors/eigenfunctions. We have
\begin{equation} \label{eq:lls-eigen}
%\begin{split}
H = \sum {\lambda_i \be_i \be_i^T}, ~~~~ \langle \be_i, \be_j \rangle = \delta_{ij}
%\end{split}
\end{equation}   
% Noticing that $\bb=H\balpha^*$ and writing the gradient in the basis of eigenvectors we obtain 
% \begin{equation} \label{eq:gradient-eigen}
% H \balpha - \bb  = H (\balpha - \balpha^*)
% = \sum_{i=1}^{d} {\lambda_i \langle \be_i, \balpha - \balpha^* \rangle \be_i}
% \end{equation}
Writing Eq.~\ref{eq:operator_grad_descent} in terms of eigendirection yields
\begin{equation} \label{eq:gd-update-rate}
\balpha^{(t+1)} - \balpha^* = \sum {(1- \eta \lambda_i)^t \langle \be_i, \balpha^{(1)} - \balpha^* \rangle \be_i}
\end{equation}
and hence, putting $a_i = \langle \be_i, \bv \rangle$, 
\begin{equation}\label{eq:reach}
\mathcal{CR}_t(\epsilon)=\{\bv, s.t. \sum {(1- \eta \lambda_i)^{2t} a_i^2 }<\epsilon^2 \sum a_i^2
\end{equation}
 
Recalling that $\eta < 2\lambda_1$ and using the fact that $(1-1/z)^z \approx 1/e$,   we  see that a necessary condition for  $\bv \in \mathcal{CR}_t$
\begin{equation}\label{eq:reach1}
  \frac{1}{3} \sum_{i, s.t. \lambda_i < \frac{\lambda_1}{2t}}  a_i^2 < \sum_i {(1- \eta \lambda_i)^{2t} a_i^2 } <\epsilon^2 \sum a_i^2
\end{equation}
This is a convenient characterization, we will denote 
\begin{equation}\label{eq:reach2}
\mathcal{CR'}_t(\epsilon) \defeq \{\bv, s.t.\sum_{i, s.t. \lambda_i <\frac{\lambda_1}{2t}}  a_i^2 < \epsilon^2\ \|\bv\|^2 \supset \mathcal{CR}_t(\epsilon) 
\end{equation}

Another  convenient  necessary condition for  $\bv \in \mathcal{CR}_t$, is that  
$$\forall_i ~~~\left|{\left(1- \frac{2\lambda_i} {\lambda_1}\right)^{t} \langle \be_i, \bv \rangle}\right| < \epsilon \|\bv\|.$$ 
Applying logarithm and noting that $\log (1-x) <-x$ results in the following inequality that must hold for all $i$ (assuming $\lambda_i<\lambda_1/2$):
\begin{equation}\label{eq:gd_number_iterations_i}
t >\frac{\lambda_1}{2\lambda_i} \log \left(\frac{ |\langle \be_i, \bv \rangle|}{\epsilon \|\bv\|}\right) 
\end{equation}

\begin{remark}
%{\bf Remark}[The condition number]: 
The standard result (see, e.g.,~\cite{boyd2004convex}) that the number of iterations necessary for uniform convergence is of the order of the condition number $\lambda_1/\lambda_d$ follows immediately. However, we are primarily interested in the case when $d$ is infinite or very large.
The corresponding operators/matrices are extremely ill-conditioned with infinite or approaching infinity condition number. In that case instead of a single condition number, one should consider a sequence of ``condition numbers'' along each eigen-direction. %\newtxtblock
\end{remark}

\subsection{Gradient descent, smoothness,  and kernel methods.}
We  now proceed to analyze the computational reach for kernel methods. We will start by discussing the case of {\it infinite data} (the population case). It is both easier to analyze and allows us to demonstrate the purely computational (non-statistical) nature of limitations of gradient descent. 

We will show that when the kernel is smooth, the reach of gradient descent is limited to very smooth, at least infinitely differentiable functions. Moreover,  to approximate a function with less smoothness within some accuracy $\epsilon$ in the $L^2$ norm one needs a super-polynomial (or even exponential) in $1/\epsilon$ number of iterations of gradient descent. 

Let the data be sampled from a probability with density $\mu$  on a compact domain $\Omega\subset \mathbb{R}^p$. 
In the case of infinite data $\H$ becomes an integral operator corresponding to a positive definite kernel $k(\cdot,\cdot)$. We have
\begin{equation}\label{eq:cov-op}
\H f (x) = \int_\Omega k(x,z) f(z) d\mu_z
\end{equation}
This is a compact self-adjoint operator with an infinite positive spectrum $\lambda_1,\lambda_2,\ldots$ with $\lim_{i \rightarrow \infty} \lambda_i=0$. 
%We will assume that $\mu$ is the uniform measure on $\mathbb{R}^p$. The case of arbitrary measure with a smooth strictly positive density can be analyzed similarly.
%(see the discussion in  Appendix~\ref{ap:eigendecay}). 

We start by stating some results on the decay of eigenvalues of $\H$. 
\begin{theorem}\label{theorem:eig_decay}
If $k$ is an infinitely differentiable kernel, the rate of eigenvalue decay is super-polynomial, i.e. 
$$
\lambda_i = O(i^{-P}) ~~~~~\forall P\in \mathbb{N}
$$
Moreover, if $k$ is an infinitely differentiable radial  kernel (e.g., a Gaussian kernel), there exist constants $C,C'>0$ such that for large enough $i$,
$$ 
\lambda_i < C'\exp\left(-C{i^{1/p} } \right) 
$$
\end{theorem}

\begin{proof} The statement for arbitrary smooth kernels is an immediate corollary of Theorem~4 in~\cite{kuhn1987eigenvalues}. 
The rate for the practically important smooth radial kernels, including Gaussian kernels, Cauchy kernel and a number of other kernel families, is given in in~\cite{Santin2016},  Theorem~6.
% is known in some special cases when the eigenvalues can be written explicitly (e.g.,~\cite{fasshauer2012stable}), but we were not able to find a general result in the literature. See Appendix~\ref{ap:eigendecay} for the proof. 
\end{proof}

\begin{remark}
Interestingly,  while eigenvalue decay is nearly exponential, it becomes milder as the dimension increases, leading to an unexpected ``blessing of dimensionality" for gradient-descent type methods in high dimension. On the other hand, while not reflected in Theorem~\ref{theorem:eig_decay}, this depends on the {\it intrinsic dimension} of the data, moderating the effect. 
\end{remark}
%Ineq.~\ref{eq:gd_number_iterations_i} implies that $t$ iterations of gradient descent guarantee convergence along at most $t^{1/P}$ eigenfunctions. \newtxtblock

\noindent {\bf The computational reach of gradient descent in kernel methods.}
Consider now the eigenfunctions of $\H$, $\H e_i = \lambda_i e_i$, which form  an orthonormal basis for $L^2(\Omega)$ by the Mercer's theorem. We can write a function $f \in L^2(\Omega)$ as $f = \sum_{i=1}^\infty a_i e_i$. We have $\|f\|_{L^2}^2 =  \sum_{i=1}^\infty a_i^2$. 

We can now  describe the reach of kernel methods with smooth kernel (in the infinite data setting). 
Specifically, functions which can be approximated in a polynomial number of iterations must have super-polynomial coefficient decay in the basis of kernel eigenfunctions. 
\begin{theorem}~\label{theorem:inapprox}
Suppose $f \in L^2(\Omega)$ is such that it can be approximated within $\epsilon$  using a polynomial in $1/\epsilon$ number of gradient descent iterations, i.e., $\forall_{\epsilon >0} f \in \mathcal{CR}_{\epsilon^{-M}}(\epsilon) $ for some $M \in \mathbb{N}$. Then for any $N \in \mathbb{N}$ and $i$ large enough $|a_i|<i^{-N}$. 
\end{theorem}
\begin{proof}
Note that $f \in {CR'}_t(\epsilon)$. We have
$
1/3\sum_{i, s.t. \lambda_i <\frac{\lambda_1 }{2\epsilon^M}}  a_i^2 < \epsilon^2\|f\|$ and hence $|a_i|< \sqrt{3} \epsilon \|f\|$ for any $i$, such that $\lambda_i <\frac{\lambda_1 }{2\epsilon^M}$.  
From  Theorem~\ref{theorem:eig_decay} we have $\lambda_i = o(i^{-NM})$. Thus this inequality holds whenever $i>C\epsilon^{-1/N}$ for some  $C$.  Writing $\epsilon$ in terms of $i$ yields $|a_i|<i^{-N}$ for  $i$ sufficiently large. 
\end{proof}
% Differentiating the kernel and using  standard convergence results for the derivative of eigenfunctions in $L^2(\Omega)$ (see Appendix~\ref{ap:concentration}), we can obtain the following
It is easy to see that the eigenfunctions $e_i$ corresponding to an infinitely differentiable kernel are also infinitely differentiable. Suppose in addition that their derivatives  grow at most polynomially in $i$, i.e. $\| e_i\|_{W_{k,2}} <P_k(i)$, where $P_k$ is some polynomial and $W_{k,2}$ is a Sobolev space. Then by differentiating the expansion of $f$ in terms of eigenfunctions we have the following 
\begin{corollary}~\label{cor:inapprox}
Any $f \in L^2(\Omega)$ that for any $\epsilon>0$ can be $\epsilon$-approximated with polynomial in $1/\epsilon$ number of steps of gradient descent is infinitely differentiable. Thus, if   $f$  is not infinitely differentiable it cannot be $\epsilon$-approximated in $L^2(\Omega)$ using a polynomial number of gradient descent steps.
\end{corollary}
%We omit the proof which can be obtained by differentiating the kernel. 
We contrast Theorem~\ref{theorem:inapprox} showing extremely slow convergence of gradient descent  with the  analysis of gradient descent for convex objective functions.
The standard analysis (e.g.~\cite{bubeck2015convex})  indicates that $O(1/\epsilon)$ steps of gradient descent is sufficient to recover the optimal value with $\epsilon$ accuracy
and at first glance seems to apply in general to infinite dimensional Hilbert spaces. 
However in this case the standard analysis cannot be used is  that the sequence $\balpha^{(t)}$ {\it diverges} in $L^2(\Omega)$ as the optimal 
%parameter vector $\balpha^*=\H^{-1}f$
solution $f^* = \H^{-1} \mathrm{y}$
is not generally a function\footnote{It can be viewed as a generalized function in the space $\H^{-1}L^2(\Omega)$.} in $L^2(\Omega)$.  This is a consequence of the fact that the infinite-dimensional  operator $\H^{-1}$ is unbounded (see Appendix~\ref{ap:weight} for some experimental results).

\begin{remark}
Note that in finite dimension every non-degenerate operator, e.g. an inverse of a kernel matrix, is bounded. Nevertheless, infinite dimensional unboundedness manifests itself as the norm of the optimal solution can increase very rapidly with the size of the kernel matrix. 
\end{remark}

\noindent {\bf Gradient descent for periodic functions on $\mathbb{R}$.} Let us now consider a simple but important special case, where gradient descent and its reach can be analyzed very explicitly. Let $\Omega$ be a circle with the uniform measure, or, equivalently, consider periodic functions on the interval $[0,2\pi]$. Let $k_s(x,z)$ be the heat kernel on the circle~\cite{rosenberg1997laplacian}. This kernel is very close to the Gaussian kernel $k_s(x,z) \approx \frac{1}{\sqrt{2 \pi s} }\exp{\left(-\frac{(x-z)^2}{4s}\right)}$.  
The eigenfunctions $e_j$ of the integral operator $\H$ corresponding to $k_s(x,z)$ are simply the Fourier harmonics\footnote{We use $j$ for the index to avoid confusion with the complex number  $i$.} $\sin j x$ and $\cos jx$. The corresponding eigenvalues are  $\{1,e^{-s}, e^{-s}, e^{-4s}, e^{-4s}, \ldots, e^{-\floor{j/2+1}^2s}, \ldots\}$ and the kernel can be written as 
$$
k_s(x,z) = \sum_0^\infty e^{-\floor{j/2+1}^2s} e_j(x) e_j(z).
$$
Given a function $f$ on $[0,2\pi]$, we can write its Fourier series 
$f = \sum_{j=0}^\infty a_j e_j$. 
A  direct computation shows that for any $f \in {CR}_t(\epsilon)$, we have $\sum_{i>\frac{\sqrt{2 \ln{2t}}}{s}}  a_i^2 < 3 \epsilon^2 \sum_0^\infty a_i^2$. We see that the space $f \in {CR}_t(\epsilon)$ is ``frozen'' as ${\sqrt{2 \ln{2t}}{s}}$ grows extremely slowly when the number of iterations $t$ increases.  
As a simple example consider the Heaviside step function $f(x)$ (on a circle),  taking $1$ and $-1$ values for $x \in (0,\pi]$ and $x\in (\pi, 2\pi]$, respectively.  The step function can be written as $f(x) = \frac{4}{\pi} \sum_{j=1,3,\ldots} \frac{1}{j}\sin(jx)$. 
From the analysis above, we need $O(\exp(\frac{s}{\epsilon^2}))$ iterations of gradient descent to obtain an $\epsilon$-approximation to the function. 
It is important to note that the Heaviside function is a rather natural example in the classification setting, where it represents the simplest two-class classification problem. 

\begin{wrapfigure}{r}{0.45\textwidth}
  %\vspace{-2mm}
  \centering
  \includegraphics[width=0.45\textwidth]{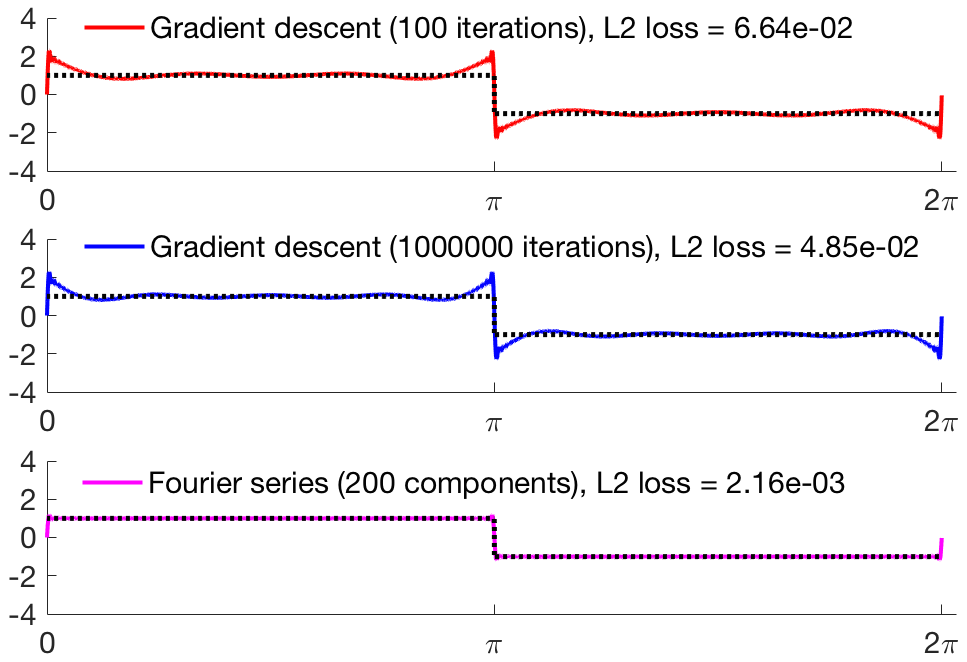}
  \vspace{-7mm}
  \caption{Top: Heaviside step function approximated by $100$ iterations of gradient descent with the Heat kernel (s=0.5). Middle: Approximation after  $10^6$ iterations of gradient descent. Bottom: Fourier series approximation with $200$ Fourier harmonics. 
}
  \label{fig:heaviside}
  \vspace{-3mm}
\end{wrapfigure}
In contrast, a direct computation of inner products $\langle f,e_i\rangle$ yields {\it exact} function recovery for any function in $L^2([0,2\pi])$ using the amount of computation equivalent to just {\it one step\footnote{Applying an integral operator, i.e. infinite dimensional matrix multiplication, is roughly equivalent to a countable number of  inner products}} of gradient descent\footnote{Of course, direct computation of inner products requires knowing the basis explicitly and in advance. In higher dimensions it also incurs a cost exponential in the dimension of the space.}.
%\footnote{cf. Nyquist-Shannon sampling theorem~\cite{shannon1949communication}}.
Thus, we see that the gradient descent is  an extremely inefficient way to recover Fourier series for a general periodic function.   
 See Figure~\ref{fig:heaviside} for an illustration of this phenomenon. We see that the approximation for the Heaviside function is only marginally improved by going from $100$ to $10^6$ iterations of gradient descent. On the other hand, just $200$ Fourier harmonics provide a far more accurate reconstruction. 

Things are not much better for  functions with more smoothness   unless they happen to be extremely smooth with exponential Fourier component decay. Thus in the classification case we expect nearly exponential increase in computational requirements as the margin between classes decreases.

% Guaranteeing $\epsilon$-approximation along the $j$th harmonic\footnote{In fact, harmonics come in pairs.} requires at least
% \begin{equation}
% \label{eq:number_iterations_Fourier}
% t>\frac{1}{2} e^{s\floor{j/2}^2} \log \left(\frac{|a_j|}{\epsilon}\right)
% \end{equation}
% iterations of gradient descent. 
% Thus after $t$ iterations we can at best estimate a function with at most 
% $O^*(\sqrt{\frac{\log t}{s}})$ Fourier components (without making assumptions about the decay of the coefficients $a_j$). 
% This is extremely inefficient as  an exponentially large number of iterations  is required to the recovery of Fourier  components. In particular, it is easy to see that for any Sobolev space (characterized by the polynomial  decay of the coefficients $a_j$, e.g.~\cite{adams2003sobolev})  on the interval $[0,2\pi]$, there is a function with norm $1$, requiring an arbitrarily large number of steps of gradient descent to approximate. 

The situation is only mildly improved in  dimension $d$, where  the span of at most $O^*\left(({\log t})^{d/2}\right)$ eigenfunctions of a Gaussian kernel or $O\left(t^{1/p}\right)$ eigenfunctions of an arbitrary $p$-differentiable kernel can be approximated in $t$ iterations.
The discussion above shows that the gradient descent with a smooth kernel can be viewed  as a  heavy regularization of the target function. It is essentially a band-limited approximation with $(\ln t)^\alpha$ Fourier harmonics for some $\alpha$. 
While regularization is often desirable from a  generalization/finite sample point of view in machine learning, especially when the number of data points is small,  the bias resulting from the application of the gradient descent algorithm cannot be overcome in a realistic number of iterations unless the target functions are extremely smooth or the kernel itself is not infinitely differentiable. %\newtxtblock

\begin{remark}[{\bf Rate of convergence vs statistical fit}]
Note that we can  improve convergence by changing the shape parameter of the kernel, i.e. making it more ``peaked'' (e.g., decreasing the bandwidth $s$ in the definition of the Gaussian kernel) While that does not change the exponential nature of the asymptotics of the eigenvalues, it slows their decay.
Unfortunately improved convergence comes at the price of overfitting. In particular, for finite data, using a very narrow Gaussian kernel results in an approximation to the $1$-NN classifier, a suboptimal method which is up to a factor of two inferior to the Bayes optimal classifier in the binary classification case asymptotically. See Appendix~\ref{ap:bandwidth} for some empirical results on the bandwidth selection for Gaussian kernels. Another possibility is to use a kernel, such as the Laplace kernel, which is not differentiable at zero. However, it also seems to consistently under-perform more smooth kernels on real data, see Appendix~\ref{ap:kernel-select} for some experiments.
\end{remark}

\noindent {\bf Finite sample effects, regularization and early stopping.} 
So far we have discussed the effects of the infinite-data version of gradient descent. We will now  discuss issues related to the finite sample setting we encounter in practical machine learning. It is well known (e.g.,~\cite{braun2005spectral, rosasco2010learning}) that the top eigenvalues of kernel matrices approximate the eigenvalues of the underlying integral operators.  Therefore computational obstructions 
encountered in the infinite case persist  whenever the data set is large enough.

Note that  for a kernel method, $t$  iterations of gradient descent for $n$ data points require $t \cdot n^2$ operations. Thus, gradient descent is computationally pointless unless $t\ll n$.  That would allow us to fit only about $O(\log t)$ eigenvectors.
In practice we would like to have  $t$ to be 
much smaller than $n$, probably a reasonably small constant.
%~\footnote{Note that SGD does not significantly change this calculus, as an epoch is roughly equivalent to one step of gradient descent, at least in our setting.}, say, $t<1000$.
%\newtxtblock

At this point we should contrast our conclusions with the important analysis  of early stopping for gradient descent provided 
in~\cite{yao2007early} (see also~\cite{raskutti2014early, camoriano2016nytro}).
The authors analyze gradient descent for kernel methods obtaining the optimal number of iterations of the form $t = n^\theta, \theta\in (0,1)$. That seems to contradict our conclusion that a very large, potentially exponential, number  of iterations may be needed to guarantee convergence. 
The apparent contradiction stems from the assumption in~\cite{yao2007early} and other works that the regression function $f^*$ belongs to the range of some power of the kernel operator $K$. For an infinitely differentiable kernel, that implies super-polynomial spectral decay ($a_i = O(\lambda_i^N)$ for any $N>0$).
In particular, it implies that $f^*$ belongs to any Sobolev space. 
We do not typically expect such high degree of smoothness in practice, particularly in classification problems. In general,   we expect sharp transitions of label probabilities across class boundaries to be typical for many classifications datasets. The Heaviside step function seems to be a simple but  reasonable model for that behavior in one dimension. 
These areas of near-discontinuity\footnote{Interestingly these sharp transitions can lead to  lower  sample complexity for optimal classifiers (cf. Tsybakov margin condition~\cite{tsybakov2004optimal}).} will result in slow decay of Fourier coefficients of $f^*$  and a mismatch with any infinitely differentiable kernel. Thus a reasonable approximation of $f^*$ would require a large number of gradient descent iterations.
\begin{wraptable}{r}{0.6\linewidth}
\vspace{-3mm}
\centering
%\caption{Gradient descent (Pegasos) on MNIST and HINT-M subset ($10000$ data points)}
%\label{tbl:gd}
\begin{adjustbox}{center}
\resizebox{9cm}{!}{
\begin{tabular}{|c|c|c||c|c|c|c|c|}
\hline
\multirow{2}{*}{Dataset} & \multicolumn{2}{c||}{\multirow{2}{*}{Metric}} & \multicolumn{5}{c|}{Number of iterations} \\ \cline{4-8} 
 & \multicolumn{2}{c||}{} & 1 & 80 & 1280 & 10240 & 81920 \\ \hline
\multirow{3}{*}{MNIST-10k} & \multirow{2}{*}{L2 loss} & train & 4.07e-1 & 9.61e-2 & 2.60e-2 & 2.36e-3 & 2.17e-5 \\ \cline{3-8} 
 &  & test & 4.07e-1 & 9.74e-2 & 4.59e-2 & 3.64e-2 & \textbf{3.55e-2} \\ \cline{2-8} 
 & \multicolumn{2}{c||}{c-error (test)} & 38.50\% & 7.60\% & 3.26\% & \textbf{2.39\%} & 2.49\% \\ \hline
\hline
\multirow{2}{*}{HINT-M-10k} & \multirow{2}{*}{L2 loss} & train & 8.25e-2 & 4.58e-2 & 3.08e-2 & 1.83e-2 & 4.21e-3 \\ \cline{3-8} 
 &  & test & 7.98e-2 & 4.24e-2 & 3.34e-2 & \textbf{3.14e-2} & 3.42e-2 \\ \hline
\end{tabular}
} \end{adjustbox}
\vspace{-3mm}
\end{wraptable}
To illustrate this point with a real data example, consider the results 
%in~\Cref{tbl:gd}.
in the table on the right.
We show the results of gradient descent for two subsets of $10000$ points from the MNIST and HINT-M datasets (see Section~\ref{sec:expr} for the description) respectively.
We see that the regression error on the training set is roughly inverse to the number of iterations, i.e. every extra bit of precision requires twice the number of iterations for the previous bit. For comparison, as  we are primarily interested in the generalization properties of the solution, we see that the minimum regression ($L^2$) error on both test sets is achieved at over $10000$ iterations. This results in at least {\it cubic} computational complexity equivalent to that of a direct method. While HINT-M is a regression dataset,  the optimal {\it classification} accuracy for MNIST is also achieved at about $10000$ iterations.

\noindent{\bf Regularization ``by impatience''/explicit regularization terms.} The above discussion suggests that gradient descent applied to kernel methods would typically result in underfitting for most larger datasets. Indeed, even $10000$ iterations of gradient descent is prohibitive when data size is more than $10^6$. As we will see in the experimental results section this is indeed the case. SGD ameliorates the problem mildly by allowing us to take approximate steps much faster but even so running standard gradient descent methods to optimality is often impractical. 
In most cases we observe little need for explicit early stopping rules. Regularization is a result of  computational constraints (cf.~\cite{yao2007early, raskutti2014early, camoriano2016nytro}) and can be termed regularization ``by impatience'' as we run out of time/computational budget allotted to the task.

%\end{remark}

% This is borne in  practice. As we will see in the experimental section, for standard gradient descent methods, the error rate on the training set decreases very slowly (Table~\ref{tbl:mnist-gd}). 
% Note that this is not a limitation of the shallow architecture as such (as the results can be dramatically improved by preconditioning) but a result of over-regularization by gradient descent. On the positive side, there is no need for extra regularization terms and parameters. Regularization is purely computational (cf.~\cite{yao2007early, raskutti2014early, camoriano2016nytro}) and can be termed regularization ``by impatience'' as we run out of time/computational budget allotted to the task.

%\begin{remark}[The futility of regularization]
Note that typical forms of regularization, result a large bias along eigenvectors with small eigenvalues $\lambda_i$. 
For example, adding  a term of the form $\lambda \|f\|_\H$ (Tikhonov regularization/ridge regression) replaces
 $\frac{1}{\lambda_i}$ by $\frac{1}{\lambda + \lambda_i}$.
While this improves the condition number and hence the speed of convergence, it comes at a high cost in terms of  over-regularization/under-fitting as it essentially discards information along eigenvectors with eigenvalues smaller than $\lambda$. 
In the Fourier series analysis example, introducing $\lambda$  this is similar to 
considering band-limited functions with $\sim\frac{\sqrt{\log(1/\lambda)}}{s}$ Fourier components. Even for $\lambda=10^{-16}$ (machine precision for double floats) and the kernel parameter $s=1$  we can only fit about $10$ Fourier components!
We argue that in most cases there is little need for explicit regularization  in the big data regimes as our primary concern is {\it underfitting}. 
%\end{remark}

\begin{remark}[Stochastic gradient descent]
%{\bf Remark: stochastic gradient descent.}
Our discussion so far has been centered entirely on gradient descent. In practice stochastic gradient descent is often used for large data.  In our setting, for fixed $\eta$, using SGD results in the same expected step size in each eigendirection as gradient descent. Hence, using SGD does not expand the algorithmic reach of gradient descent, although it  speeds up convergence in practice. On the other hand, SGD introduces a number of interesting algorithmic and statistical subtleties. We will address some of them below. 

\end{remark}

\section{EigenPro iteration: extending the reach of gradient descent }\label{sec:linear}
%%%%%
% The approach to address slow convergence of gradient descent taken in this work is to precondition the covariance matrix $H$. We will call the resulting algorithm  EigenPro,  
% a systematic approach to preconditioning least square regression problems. 
% %a class of such preconditioned methods.

% To define the preconditioned gradient descent, let's first modify the linear system in Eq.~\ref{eq:linear} with an invertible matrix $P$, called a left preconditioner.
% \begin{equation}
% \label{eq:pc-linear}
% PH\alpha - Pb = 0
% \end{equation}
% Clearly, the modified system in Eq.~\ref{eq:pc-linear} and the original system in Eq.~\ref{eq:linear} have the same solution. Now consider the Richardson iteration corresponding to the modified system,
% \begin{equation}
% \label{eq:pc-gd}
% \alpha^{(t+1)} = \alpha^{(t)} - \eta P (H \alpha^{(t)} - b)
% \end{equation}
% The modified iteration is a preconditioned gradient descent method. It converges to $\alpha^*$, the solution of the original linear system.  %The matrix $P$ is called a left preconditioner.

We will now propose some practical measures to alleviate the issues related to over-regularization of  linear regression by gradient descent. 
As seen above, one of the key shortcomings of shallow learning methods based on smooth kernels (and their approximations, e.g. Fourier and RBF features) is their fast spectral decay.  That observation suggests  modifying the corresponding matrix $H$ by decreasing its top eigenvalues. This ``partial whitening'' enables the algorithm to approximate more target functions in a fixed number of iterations. 

It turns out that accurate  approximations of the top eigenvectors can be obtained from small subsamples of the data with modest computational expenditure. 
Moreover,  ``partially whitened" iteration can be done in a way compatible with stochastic gradient descent thus obviating the need to materialize full covariance/kernel matrices in memory. Combining these observations we construct a low overhead preconditioned Richardson iteration which we call EigenPro iteration. \newtxtblock

\noindent {\bf Preconditioned (stochastic) gradient descent.} 
%We will call the resulting algorithm  EigenPro, a systematic approach to preconditioning least square regression problems. 
%Recall that preconditioning  consists in modifying 
We will modify the linear system in Eq.~\ref{eq:linear} with an invertible matrix $P$, called a left preconditioner.
\begin{equation}
\label{eq:pc-linear}
PH\balpha - P\bb = 0
\end{equation}
Clearly, the modified system in Eq.~\ref{eq:pc-linear} and the original system in Eq.~\ref{eq:linear} have the same solution.  The Richardson iteration corresponding to the modified system (preconditioned Richardson iteration) is
\begin{equation}
\label{eq:pc-gd}
\balpha \leftarrow \balpha - \eta P (H \balpha - \bb)
\end{equation}
It is easy to see that as long as $\eta \|P H\| < 1$ it converges to $\balpha^*$, the solution of the original linear system.

Preconditioned SGD can be defined  similarly by
\begin{equation}
\label{eq:pc-sgd}
\balpha \leftarrow \balpha - \eta P (H_m \balpha - \bb_m)
\end{equation}
where we define $H_m \defeq \frac{2}{m} X_m^T X_m$ and $b_m \defeq \frac{2}{m} X^T_m \by_m$ using $(X_m, \by_m)$, a sampled mini-batch of size $m$.
This preconditioned iteration also converges to $\balpha^*$ with properly chosen $\eta$~\cite{murata1998statistical}.

\begin{remark}
Notice that the preconditioned covariance matrix $PH$ does not in general have to be  symmetric. 
 It is sometimes convenient to consider the closely related iteration
 \begin{equation}\label{eq:lrpc-richardson}
 \bbeta \leftarrow \bbeta
- \eta (P^\frac{1}{2}H P^\frac{1}{2} \bbeta  - P^\frac{1}{2} \bb)
\end{equation}
Here $P^\frac{1}{2}H P^\frac{1}{2}$ is a symmetric matrix. We see that $\bbeta^* = P^{-1/2}\balpha^*$. 
\end{remark}

\noindent {\bf Preconditioning as a linear feature map.} It is easy to see that  preconditioned iteration in Eq.~\ref{eq:lrpc-richardson} is in fact equivalent to the standard Richardson iteration in Eq.~\ref{eq:richardson} on a dataset transformed with the linear feature map,
\begin{equation}\label{eq:linear-fm}
\phi_P(\bx) \defeq P^\frac{1}{2} \bx 
\end{equation}
This is a convenient point of view as the transformed data can be stored for future use. It also shows that preconditioning is compatible with most computational methods both in practice and, potentially, in terms of analysis. \newtxtblock

% To see the connection, rewrite Eq.~\ref{eq:lrpc-richardson} as
% \begin{equation}\label{eq:expand-lrpc-richardson}
% \bbeta \leftarrow \bbeta
% - \eta (\frac{1}{n} \Phi^T \Phi \bbeta - \Phi^T \by)
% \end{equation}
% where $\Phi \defeq (\phi_P(\bx_1), \ldots, \phi_P(\bx_n))^T = X P^\frac{1}{2}$ is the transformed dataset. If we generate $\Phi$ in advance, Eq.~\ref{eq:expand-lrpc-richardson} is exactly Eq.~\ref{eq:richardson}, the standard GD.
%Thus, another option to perform preconditioning is to preprocess the dataset by $P^\frac{1}{2}$, then train the linear model defined in Eq.~\ref{eq:linear} on the transformed dataset $\Phi$. 
%However, when the features of the data are generated online and the number of them is large, this offline version preconditioning is impractical due to the limitation of memory space.

%This is a typical setting encountered in applications of kernel methods to large modern data, where the dimension of the feature space grows with the size of the data. 

\subsection{Linear EigenPro}
We will now discuss properties desired to make preconditioned GD/SGD methods effective on large scale problems.
Thus for the modified iteration in Eq.~\ref{eq:pc-gd} we would like to choose $P$ to meet the following targets:

\noindent {\bf Acceleration.} The algorithm should provide high accuracy
in a small number of iterations.

\noindent {\bf Initial cost.} The preconditioning matrix $P$ should be  accurately computable without materializing the  full covariance matrix.

\noindent {\bf Cost per iteration.} Preconditioning by $P$ should be efficient per iteration in terms of computation and memory. \newtxtblock

%%%%%%%% Linear EigenPro (E) %%%%%%%%
%\noindent {\bf Linear EigenPro.}

% Note that when $\lambda_i \ll \lambda_i$, the convergence of the preconditioned  algorithm  %{\it optimal step size} $\eta$ 
% along the $i$-th eigendirection after $t$ iterations depends on the ratio of  $\left(1-\frac{\lambda_i(\PH)}{\lambda_1(\PH)}\right)^t\approx 1-t\frac{\lambda_i(\PH)}{\lambda_1(\PH)}$, 
% Specifically,
% %according to Eq.~\ref{eq:relative-approx-error},
\begin{wraptable}{r}{0.5\linewidth}
\vspace{-4mm}
\begin{subalgorithm}{\linewidth}
\begin{algorithmic}[1]
\ALG $\mathrm{EigenPro}(X, \by, k, m, \eta, \tau, M)$
  \INPUT
	training data $(X, \by)$,
	number of eigen-directions $k$,
	mini-batch size $m$,
    step size $\eta$,
	damping factor $\tau$,
    subsample size $M$
  \OUTPUT
	weight of the linear model $\balpha$
  	\STATE $[E, \Lambda, \hat{\lambda}_{k+1}] = \mathrm{RSVD}(X, k+1, M)$
    \STATE $P \defeq I - E (I - \tau \hat{\lambda}_{k+1} \Lambda^{-1}) E^T$
    \STATE Initialize $\balpha \leftarrow 0$
  	\WHILE{stopping criteria is False}
    	\STATE $(X_m, \by_m) \leftarrow$ $m$ rows sampled from $(X, \by)$
        without replacement
        \STATE $\bg \leftarrow \frac{1}{m}(X_m^T (X_m \balpha) - X_m^T \by_m)$
%       \STATE $\bg \leftarrow \frac{1}{m}(\frac{1}{d}X_m^T (X_m \balpha) - \frac{1}{\sqrt{d}} X_m^T \by_m)$
  		\STATE $\balpha \leftarrow \balpha - \eta P \bg$
    \ENDWHILE
\end{algorithmic}
\end{subalgorithm}
\captionsetup{labelformat=alglabel}
\vspace{1mm}
\caption{EigenPro iteration in vector space}%
\label{alg:epro}
\vspace{-7mm}
\end{wraptable}
The relative approximation error along $i$ the eigenvector for gradient descent  after $t$ iterations  is $\left(1-\frac{\lambda_i(\PH)} {\lambda_1(\PH)}\right)^t$. % with optimal learning rate $\frac{1}{\lambda_1(\PH)}$.
Minimizing the error suggests choosing the preconditioner $P$  to maximize the ratio $\frac{\lambda_i(\PH)} {\lambda_1(\PH)}$ for each $i$. 
% is directly dependent on the learning rate parameter 
% According to Eq.~\ref{eq:relative-approx-error}, the relative approximation error along eigen-direction $e_i$ for GD at iteration $t+1$ is $e^{-t \frac{\lambda_i(H)} {\lambda_1(H)}}$ with optimal learning rate $\frac{1}{\lambda_1(H)}$. Therefore, smaller $\frac{\lambda_i(H)} {\lambda_1(H)}$ leads to faster convergence of GD. If seeing preconditioning as a linear feature map, the convergence of the preconditioned GD and SGD relies on ratios, $\frac{\lambda_i(\PH)} {\lambda_1(\PH)}$ and $\frac{\lambda_i(\PH_m)} {\lambda_1(\PH_m)}$, respectively. Thus, to achieve high acceleration with preconditioning, we want such ratio as small as possible.
We see that modifying the top eigenvalues of $H$ makes the most difference in convergence. For example, decreasing $\lambda_1$ improves convergence along all directions, while decreasing any other eigenvalue only speeds up convergence in that direction . However, decreasing $\lambda_1$ below $\lambda_2$ does not help unless $\lambda_2$ is decreased as well. Therefore it is natural to decrease the top $k$ eigenvalues to the maximum amount, i.e. to $\lambda_{k+1}$, leading to the preconditioner
\begin{equation} \label{eq:epro-pc}
\begin{split}
P \defeq I - \sum_{i=1}^k {(1 - \frac{\lambda_{k+1}} {\lambda_i}) \be_i \be_i^T}
\end{split}
\end{equation}
In fact it can be readily seen that $P$ is {\it the optimal preconditioner} of the form $I - Q$, where $Q$ is a low rank matrix.
We will see that $P$-preconditioned iteration accelerates convergence by approximately a factor of $\lambda_1/\lambda_k$.  

%%%%%%%%%%%%%%%%
However, exact construction of $P$ involves computing the eigendecomposition of the $d \times d$ matrix $H$, which is not feasible for large data size.  To avoid this, we use subsample randomized SVD~\cite{halko2011finding} to obtain an approximate preconditioner, defined as
\begin{equation} \label{eq:approx-epro-pc}
\begin{split}
\hat{P}_{\tau} \defeq I - \sum_{i=1}^k {(1 - \tau \frac{\hat{\lambda}_{k+1}} {\hat{\lambda}_i}) \hat{\be}_i \hat{\be}_i^T}
\end{split}
\end{equation}
where
algorithm RSVD (see Appendix~\ref{ap:rsvd}) computes the approximate top eigenvectors $E \leftarrow (\hat{\be}_1, \ldots, \hat{\be}_k)$ and eigenvalues $\Lambda \leftarrow \textrm{diag}(\hat{\lambda}_1, \ldots, \hat{\lambda}_{k})$ and $\hat{\lambda}_{k+1}$ for subsample covariance matrix $H_M$.
%$\hat{\be}_1, \ldots, \hat{\be}_k$ and $\hat{\lambda}_1, \ldots, \hat{\lambda}_{k+1}$ are the approximate top eigenvectors and eigenvalues of $H_M$ computed by Algorithm RSVD 
% When $\tau = 1$, both definitions have the similar form. While since $\left \langle \be_i, \hat{\be}_i  \right \rangle$ decreases as $i$ increases most time, we often have $\frac{\lambda_i(PH)}{\lambda_1(PH)} > \frac{\lambda_i(\hat{P}_1 H)}{\lambda_1(\hat{P}_1 H)}$, which indicates slower convergence with approximate preconditioner.
Alternatively, a Nyström method based SVD (see Appendix~\ref{ap:rsvd}) can be applied to obtain eigenvectors (slightly less accurate although with little impact on training in practice) through a highly efficient implementation for GPU.

Additionally, we introduce the parameter $\tau$ to counter the effect of approximate top eigenvectors ``spilling'' into the span of the remaining eigensystem. Using $\tau <1$ is preferable to the obvious alternative of decreasing the step size $\eta$ as it does not decrease the step size in the directions nearly orthogonal to the span of $(\hat{\be}_1, \ldots, \hat{\be}_k)$. That allows the iteration to converge faster in those directions. In particular, $(\hat{\be}_1, \ldots, \hat{\be}_k)$ are computed exactly, the step size in other eigendirections will not be affected by the choice of $\tau$.   
% 
%derived from the adoption of approximate eigensystem. Further analysis on $\tau$ can be found in Section~\ref{sec:step}.

% In sum, using the approximate preconditioner (Eq.~\ref{eq:approx-epro-pc}) in preconditioned SGD (Eq.~\ref{eq:pc-sgd}) is the {\bf core idea of EigenPro}.
We call SGD with the preconditioner $\hat{P}_{\tau}$  (Eq.~\ref{eq:pc-sgd}) {\it  EigenPro iteration}.
% It turns out that $P$ can be approximated effectively using subsample randomized SVD and applied efficiently under the SGD setting. 
The details of the algorithm are given in Table~\ref{alg:epro}. 
Moreover, the key  step size parameter $\eta$ can be selected in a theoretically sound way discussed below.
%%%%%%%% Linear EigenPro (E) %%%%%%%%

%%%%%%%% Kernel EigenPro (S) %%%%%%%%
\subsection{Kernel EigenPro}
While EigenPro iteration can be applied to any linear regression problem,
it is particularly useful in conjunction with smooth kernels which have fast eigenvalue decay.  
We will now discuss modifications needed to work directly in the RKHS (primal) setting. %\newtxtblock

%\noindent {\bf Kernel EigenPro.}
\begin{wraptable}{r}{0.5\linewidth}
\vspace{-2mm}
\begin{subalgorithm}{\linewidth}
\begin{algorithmic}[1]
\ALG $\mathrm{EigenPro}(\rmk(\cdot, \cdot), X, \by, k, m, \eta, s_0)$
  \INPUT
    kernel function $\rmk(\cdot, \cdot)$,
    training data $(X, \by)$,
    number of eigen-directions $k$,
    mini-batch size $m$,
    step size $\eta$,
    subsample size $M$,
    damping factor $\tau$
%	\Statex $X, \by, k, m, \eta, c_0$ - same as algorithm on the left
  \OUTPUT
  	weight of the kernel method $\balpha$

  	\STATE $K \defeq \rmk(X, X)$ materialized on demand
  	\STATE $[E, \Lambda, \lambda_{k+1}] \leftarrow \mathrm{RSVD}(K, k+1, M)$
    \STATE $D \defeq E \Lambda^{-1} (I - \tau \lambda_{k+1} \Lambda^{-1})E^T$
    \STATE Initialize $\balpha \leftarrow 0$
 
  	\WHILE{stopping criteria is False}
    	\STATE $(K_m, \by_m) \leftarrow$ $m$ rows sampled from $(K, \by)$ %without replacement
        \STATE $\balpha_m \defeq$ portion of $\balpha$ related to $K_m$
        \STATE $\bg_m \leftarrow \frac{1}{m}(K_m \balpha - \by_m)$
  		\STATE $\balpha_m \leftarrow \balpha_m - \eta \bg_m$, $\balpha \leftarrow \balpha + \eta D K_m^T \bg_m$
    \ENDWHILE
\end{algorithmic}
\end{subalgorithm}
\captionsetup{labelformat=alglabel}
\vspace{1mm}
\caption{EigenPro iteration in RKHS space}%
\label{alg:primal-ep}
\vspace{-5mm}
\end{wraptable}
In this setting, a reproducing kernel $\rmk(\cdot, \cdot): \mathbb{R}^N \times \mathbb{R}^N \rightarrow \mathbb{R}$ implies a feature map from $X$ to an RKHS space $\mathcal{H}$ (typically) of infinite dimension. The feature map can be written as  $\phi:  x \mapsto k(x, \cdot), \mathbb{R}^N \to \mathcal{H}$.
This feature map leads to the (shallow) learning problem
$$
f^*=\arg\min_{f \in \mathcal{H}} \frac{1}{n} \sum_{i=1}^{n} {( \langle f, \rmk(\bx_i, \cdot) \rangle_{\mathcal{H}} - y_i)^2}
$$

Using properties of RKHS, EigenPro iteration in $\HH$ becomes $f \leftarrow f - \eta \rmP_\tau (\rmH(f)- \rmb)$
where covariance operator
$\rmH \defeq \frac{2}{n} \sum_{i=1}^n{\rmk(\bx_i, \cdot) \otimes \rmk(\bx_i, \cdot)}$ and $\rmb \defeq \frac{2}{n} \sum_{i=1}^n {y_i k(\bx_i, \cdot)}$.
The top eigensystem of $\rmH$ forms the preconditioner
$$
\rmP_\tau \defeq \rmI - \sum_{i=1}^{k}{(1 -\tau\frac{\lambda_{k+1}(\rmH)}{\lambda_i(\rmH)}) \rme_i(\rmH) \otimes \rme_i(\rmH)}
$$

Notice that by the Representer theorem~\cite{aronszajn1950theory}, $f^*$ admits a representation of the form $\sum_{i=1}^{n}{\alpha_i \rmk(\bx_i, \cdot)}$.
Parameterizing the above iteration accordingly and applying some linear algebra lead to the following iteration in a finite-dimensional vector space,
$$
\balpha \leftarrow \balpha - \eta P_\tau (K \balpha - \by)
$$ 
where $K \defeq [\rmk(\bx_i, \bx_j)]_{i,j = 1, \ldots, n}$ is the kernel matrix and EigenPro preconditioner $P$ is defined using the top eigensystem of $K$ (assume $K\be_i = \lambda_i \be_i$),
$$
P_\tau \defeq I - \sum_{i=1}^k
{\frac{1}{\lambda_i}
(1 - \tau \frac{\lambda_{k+1}}{\lambda_i}) \be_i \be_i^T}
$$

This preconditioner differs from that for the linear case (Eq.~\ref{eq:epro-pc}) with an extra factor of $\frac{1}{\lambda_i}$ due to the difference between the parameter space of $\alpha$  and the RKHS space. Table~\ref{alg:primal-ep} details the SGD version of this iteration.
\newtxtblock

\noindent {\bf EigenPro as kernel learning.}
Another way to view EigenPro is in terms of kernel learning. 
Assuming that the preconditioner is computed exactly, we see that in the population case EigenPro is equivalent to computing the (distribution-dependent) kernel 
$$
k_{EP}(x,z) \defeq \sum_{i=1}^k \lambda_{k+1} e_i(x)e_i(z) + 
\sum_{i=k+1}^\infty \lambda_{i} e_i(x)e_i(z)
$$
Notice that the RKHS spaces corresponding to $k_{EP}$ and $k$ contain the same functions but have different norms. The norm in $k_{EP}$ is a finite rank modification of the norm in the RKHS corresponding to $k$, a setting  reminiscent of~\cite{sindhwani2005beyond} where unlabeled data was used to ``warp'' the norm for semi-supervised learning.  However, in our paper the ``warping" is purely for computational efficiency.

%%%%%%%% Kernel EigenPro (E) %%%%%%%%

%%%%%%%% Cost & Acceleration (S) %%%%%%%%
\subsection{Costs and Benefits}
We will now discuss the acceleration provided by EigenPro and the overhead associated with the algorithm.

%overhead associated to the algorithm and the acceleration provided by EigenPro.
\noindent {\bf Acceleration.} 
Assuming that the preconditioner $P$ can be computed exactly, 
EigenPro computes the solution exactly in the span of the top $k+1$ eigenvectors. 
For $i>k+1$ EigenPro provides the acceleration factor of 
$
\alpha= \frac{\left(1-{\lambda_{i}}/ {\lambda_{1}}\right)^t}{\left(1-\lambda_i/ \lambda_{k+1}\right)^t}$ along the $i$th eigendirection. Assuming that $\lambda_i \ll \lambda_1$, a simple calculation shows an acceleration factor of at least   $\frac{\lambda_1}{\lambda_{k+1}}$
over the standard gradient descent. 
Note that  this assumes full gradient descent and exact computation of the preconditioner.  See below for an acceleration analysis in the SGD setting resulting in a potentially somewhat smaller acceleration factor. 

\noindent {\bf Initial cost.}
To construct the preconditioner $P$, we perform RSVD (Appendix~\ref{ap:rsvd}) to compute the approximate top eigensystem of covariance $H$. 
Algorithm RSVD has time complexity $O(Md\log{k} + (M + d) k^2)$ (see~\cite{halko2011finding}). The subsample size $M$ can be much smaller than the data size $n$ while still preserving the accuracy of estimation for top eigenvectors.
In addition, we need extra $kd$ memory to store the top-$k$ eigenvectors.

\noindent {\bf Cost per iteration.}
For standard SGD with $d$ features (or kernel centers)  and mini-batch of size $m$, the computational cost per iteration is $O(md)$. 
In addition, applying the preconditioner $P$ in  EigenPro  requires left multiplication by a matrix of rank $k$. That involves $k$ vector-vector dot products for vectors of length $d$, resulting in $k \cdot d$  operations per iteration.
Thus  EigenPro  using top-$k$ eigen-directions needs $O(md+kd)$ operations per iteration.  Note that these can be implemented efficiently on a GPU. See Section~\ref{sec:expr} for actual overhead per iteration  achieved in practice.

%%%%%%%% Cost & Acceleration (E) %%%%%%%%

\section{Step Size Selection for EigenPro Preconditioned Methods}
\label{sec:step}
We will now discuss the key issue of the step size selection for EigenPro  iteration.
For iteration involving covariance matrix $H$,  $\eta = \norm{H}^{-1}$ results in optimal (within a factor of $2$) convergence. 

This suggests choosing the  corresponding step size  $\eta = \norm{\PH}^{-1} = \lambda_{k+1}^{-1}$.
%%%%%%%%% Optimality ********
% In fact, EigenPro preconditioner $P$ is optimal in the sense that
% \begin{claim}[Optimality of EigenPro preconditioner]\label{thm:eigenpro-opt}
% For any preconditioner of form $I - D$ where $D$ is a rank-$k$ matrix, we have
% \begin{equation} \label{eq:eigenpro-opt}
% \norm{PH} \leq \norm{(I - D) H}
% \end{equation}
% \end{claim}
% Thus among all such preconditioners, EigenPro preconditioner leads to the largest feasible step size since $\norm{PH}^{-1} \geq \norm{(I - D) H}^{-1}$.
However, in practice this will lead to divergence due to (1) approximate computation of eigenvectors (2) the randomness inherent in SGD. One possibility would be to compute $\norm{PH_m}$ at every step. That, however, is  costly,  requiring computing the top singular value for every mini-batch.
% However, under the SGD setting (to achieve T1b, low cost per iteration), each iteration involves a subsample covariance matrix $H_m$ whose norm is a random variable. Although calculating $\norm{H_m}$, the top singular value of a $m \times d$ matrix, is feasible, it is a costly procedure when executing every iteration. Especially, it involves iterations unable to be parallelized on GPU.
As the mini-batch can be assumed to be chosen at random, we propose using a lower bound on $\norm{H_m}^{-1}$ (with high probability) as the step size to guarantee convergence at each iteration, which works well in practice. 
\newtxtblock

\noindent {\bf Linear EigenPro.}
Consider the EigenPro preconditioned SGD in Eq.~\ref{eq:pc-sgd}.
For this analysis assume that $P$ is formed by the exact eigenvectors\footnote{ Approximate preconditioner with $\hat{P}$ instead of $P$ can also be analyzed using results from~\cite{halko2011finding}.} of $H$.
Interpreting $P^\frac{1}{2}$ as a linear feature map as in Section~\ref{sec:preliminary}, makes $\PHmP$ a random subsample on the dataset $XP^\frac{1}{2}$. Now applying Lemma~\ref{clm:bernstein-sgd} (Appendix~\ref{ap:concentration}) results in

\begin{theorem}
\label{clm:bernstein-pc-sgd}
If $\norm{\bx}^2_2 \leq \kappa$ for any $\bx\in X$ and $\lambda_{k+1} = \lambda_{k+1}(H)$, $\norm{P H_m}$ has following upper bound with probability at least $1 - \delta$,
\begin{equation} \label{eq:bernstein-pc-sgd}
\begin{split}
\norm{PH_m}
%= \norm{\PHmP} & \\
\leq \lambda_{k+1} + \frac{2(\lambda_{k+1} + \kappa)}{3m} \ln{\frac{2d}{\delta}} 
 + \sqrt{\frac{2 \lambda_{k+1} \kappa}{m} \ln{\frac{2d}{\delta}}}
\end{split}
\end{equation}
\end{theorem}

\noindent {\bf Kernel EigenPro.} For EigenPro iteration in RKHS space, we can bound $\norm{\rmP \circ \rmH_m}$ with a similar theorem where $\rmH_m$ is the subsample covariance operator and $\rmP$ is the corresponding EigenPro preconditioner operator. Since $d = dim(\rmH)$ is infinite, we introduce {\it intrinsic dimension} from \cite{tropp2015introduction} $\mathrm{intdim}(A) \defeq \frac{tr(A)}{\norm{A}}$ where $A: \mathcal{H} \rightarrow \mathcal{H}$ is an arbitrary operator. It can be seen as a measure of the number of dimensions where $A$ has significant spectral content.
Let $\rmd \defeq \mathrm{intdim}(\mathbb{E}[(\rmH_m - \rmH) \circ (\rmH_m - \rmH)])$. Then by Lemma~\ref{clm:bernstein-sgd-op} (Appendix~\ref{ap:concentration}), we have

\begin{theorem}
\label{clm:bernstein-pc-sgd-op}
If $k(\bx, \bx) \leq \kappa$ for any $\bx\in X$ and $\lambda_{k+1} = \lambda_{k+1}(\rmH)$, with probability at least $1 - \delta$ we have,
\begin{equation} \label{eq:bernstein-pc-sgd-op}
\begin{split}
\norm{\rmP \circ \rmH_m}
\leq \lambda_{k+1} + \frac{2(\lambda_{k+1} + \kappa)}{3m} \ln{\frac{8 \rmd}{\delta}} 
 + \sqrt{\frac{2 \lambda_{k+1} \kappa}{m} \ln{\frac{8 \rmd}{\delta}}}
\end{split}
\end{equation}
\end{theorem}

\noindent {\bf Choice of the step size.} In both spectral norm bounds Eq.~\ref{eq:bernstein-pc-sgd} and Eq.~\ref{eq:bernstein-pc-sgd-op}, $\lambda_{k+1}$ is the dominant term when the mini-batch size $m$ is large. 
However, in most large-scale settings, $m$ is small, and $\sqrt{\frac{2 \lambda_{k+1}\kappa}{m}}$ becomes the dominant term.
This suggests choosing step size $\eta \sim \sqrt{\frac{m}{\lambda_{k+1}}}$ leading to acceleration on the order of $\sqrt{\frac{\lambda_1}{\lambda_{k+1}}}$ over the standard (unpreconditioned) SGD.  That choice works well in practice. 

\section{EigenPro and Related Work}

Recall that the setting of large 
scale machine learning imposes some fairly specific 
requirements on the optimization methods. In particular, 
the computational budget allocated to the problem must 
not significantly exceed $O(n^2)$ operations, i.e., a 
small  number of matrix-vector multiplications.  
That restriction rules out most direct second order methods  which require $O(n^3)$ operations. Approximate second order methods are far more effective computationally. However, they  typically rely on low rank matrix approximation, a strategy which underperforms in conjunction with smooth kernels  as information along important eigen-directions with small eigenvalues is  discarded. Similarly, regularization improves conditioning and convergence but also discards information by biasing small eigen-directions. 

While first order methods preserve important information, they, as discussed in this paper,  are too slow to converge along eigenvectors with small eigenvalues. It is clear  that an effective method must thus be a hybrid approach using approximate second order information  in a first order method. 

EigenPro is an example of such an approach as the second order information is used in conjunction with an iterative first order method. The things that make EigenPro effective are the following:\\
1. The second order information (eigenvalues and eigenvectors) is computed efficiently from a subsample of the data. Due to the quadratic loss function, that computation needs to be conducted only once. Moreover, the step size can be fixed throughout the iteration. 
\\
2. Preconditioned Richardson iteration is efficient and has a natural stochastic version. Preconditioning by a low rank modification of the identity matrix results in low overhead per iteration. The preconditioned update is computed on the fly without a need to materialize the full preconditioned covariance.\\
3. EigenPro iteration converges (mathematically) to the same result independently of the preconditioning matrix\footnote{We note, however, that convergence will be slow
if $P$ is poorly approximated.}. That makes EigenPro relatively robust to errors in the second order preconditioning term $P$, in contrast to most approximate second order methods.

We will now discuss some related literature and connections. 
\newtxtblock

% {Optimization }
% In this section we  discuss some related work in optimization, kernel methods, and linear system preconditioning. 

% It is important to reiterate that the setting of large scale machine learning presents some fairly specific requirements on the methods. In particular, 

% From the perspective of computational budget, discussion on optimization methods raises the necessity to strike a balance between first order and second order methods for shallow learning.
% Such balance is partially achieved by EigenPro iteration.
% Moreover, simple combination of EigenPro and scalable kernel methods in discussion leads to highly practical methods with state-of-the-art performance.
% We then look into alternative approaches to precondition the kernel matrix.
% \newtxtblock

\noindent {\bf First order optimization methods.}
Gradient based methods, such as gradient descent (GD), stochastic gradient descent (SGD), 
%and conjugate gradient (CG) 
are classic textbook methods~\cite{shewchuk1994introduction,dennis1996numerical,boyd2004convex,bishop2007pattern}.
%Recent renaissance of neural network drew great attentions to improving SGD, including step size adaptivity~\citep{duchi2011adaptive,kingma2014adam,tieleman2012lecture} and variance reduction gradients~\cite{johnson2013accelerating, wang2013variance}.
Recent renaissance of neural networks had drawn significant attention to improving and accelerating these methods, especially, the highly scalable mini-batch SGD.
%Related work includes step size adaptivity~\citep{duchi2011adaptive,kingma2014adam,tieleman2012lecture} and variance reduction gradients~\cite{johnson2013accelerating, wang2013variance}.
Methods like SAGA~\cite{roux2012stochastic} and SVRG~\cite{johnson2013accelerating} improve stochastic gradient by periodically evaluated full gradient to achieve variance reduction.
Another set of approaches~\cite{duchi2011adaptive, tieleman2012lecture, kingma2014adam} compute adaptive step size for each gradient coordinate every iteration. The step size is normally chosen to minimize certain regret bound of the loss function.
Most of these methods introduce affordable $O(d)$ computation and memory overhead.
%These techniques are typically designed for the DNN setting, and are generally less effective in shallow learning.

\begin{remark}
Interpreting EigenPro iteration as a linear ``partial whitening" feature map, followed by Richardson iteration, we see that most of these first order  methods are compatible with EigenPro.
Moreover,  many convergence bounds for these methods~\cite{boyd2004convex,roux2012stochastic,johnson2013accelerating} involve the condition number $\lambda_1(H) / \lambda_d(H)$. EigenPro iteration generically improves such bounds by (potentially) reducing the condition number to $\lambda_{k+1}(H) / \lambda_d(H)$.
%However, these first order techniques are typically designed for general optimization setting and do not handle fast decaying eigenvalues particularly. Thus they are usually less effective in shallow learning.
\end{remark}

\noindent {\bf Second order/hybrid optimization methods.} Second order methods use the inverse of the Hessian matrix or its approximation  to accelerate convergence \cite{ schraudolph2007stochastic,bordes2009sgd, moritz2016linearly, byrd2016stochastic,agarwal2016second}.
A limitations of many of these methods is the need to compute the full gradient instead of the stochastic gradient every iteration~\cite{liu1989limited,erdogdu2015convergence, agarwal2016second} making them harder to scale to large data.

We  note the  work~\cite{erdogdu2015convergence} which analyzed a hybrid first/second order method for general convex optimization with a rescaling term based on the top eigenvectors of the Hessian. That can be viewed as preconditioning the Hessian at every iteration of gradient descent. 
%Compared to EigenPro iteration there is a significant computational penalty resulting from recomputing the eigendecomposition at every step.
A related recent  work~\cite{gonen2016solving} analyses a hybrid method   designed to accelerate SGD convergence for linear regression with ridge regularization. The data are preconditioned  by preprocessing (rescaling) all data points along the top singular vectors of the data matrix.  The authors provide a detailed analysis of the algorithm depending on the regularization parameter.  
Another recent second order method PCG~\cite{avron2016faster} accelerates the convergence of conjugate gradient on large kernel ridge regression using a novel preconditioner.
The preconditioner is the inverse of an approximate covariance generated with random Fourier features. By controlling the number of random features, this method strikes a balance between preconditioning effect and computational cost.
\cite{tu2016large} achieves similar preconditioning effects by solving a linear system involving a subsampled kernel matrix every iteration.
While not strictly a preconditioner Nyström with gradient descent(NYTRO)~\cite{camoriano2016nytro} also improves the condition number. 
Compared to many of these methods EigenPro directly addresses the underlying issues of slow convergence without introducing a bias in directions with small eigenvalues and incurring only a small overhead per iteration both in memory and computation.

Finally, limited memory BFGS (L-BFGS)~\cite{liu1989limited} and its variants~\cite{schraudolph2007stochastic, moritz2016linearly, byrd2016stochastic} are among the most effective second order methods for unconstrained nonlinear optimization problems. 
% They typically maintain several vector pairs and apply them to the gradient through a two-loop recursion on every iteration. However, these methods incur significant overhead in a multiclass setting. 
% Specifically, when each data point has an $l$-length label vector and the methods maintain $k$ vector pairs, 
Unfortunately, they can introduce prohibitive memory and computation overhead for large multi-class problems.
\newtxtblock

\noindent {\bf Scalable kernel methods.} There is a significant literature 
on scalable kernel methods including~\cite{kivinen2004online,hsieh2008dual,shalev2011pegasos,takac2013mini,dai2014scalable}. Most of these are first order optimization methods. 
%To trade accuracy for lower computation/memory cost, they often adopt approximations like Nystr{\"o}m~\cite{queback,tu2016large,williams2001using} or random Fourier feature~\cite{rahimi2007random,tu2016large,dai2014scalable}.
To avoid the $O(n^2)$ computation and memory requirement typically involved in constructing the kernel matrix, they often adopt approximations like RBF feature~\cite{williams2001using,queback,tu2016large} or random Fourier features~\cite{rahimi2007random,dai2014scalable,tu2016large}, which reduces such requirement to $O(nd)$.
Exploiting  properties of random matrices and the Hadamard transform, \cite{le2013fastfood} further reduces the $O(nd)$ requirement to $O(n \log{d})$ computation and $O(n)$ memory, respectively.

\begin{remark}[Fourier and other feature maps]
As discussed above, most scalable kernel methods suffer from limited computational reach when used with Gaussian and other smooth kernels. Feature maps, such as Random Fourier Features~\cite{rahimi2007random}, are  non-linear transformations and are agnostic with respect to the optimization methods. Still they can be viewed as approximations of smooth kernels and thus suffer from the fast decay of eigenvalues. 
 %and are thus subject to the limitations discussed in this paper. 

% Further adoption of RBF and random Fourier feature helps scaling the method to larger dataset. Note we do not use Fastfood feature in this work due to the absence of efficient fast Hadamard transform implementation on GPU.
\end{remark}

\noindent {\bf Preconditioned linear systems.}
There is a vast literature on preconditioned linear systems with a number of recent papers
focusing on preconditioning kernel matrices, such as for low-rank approximation~\cite{fasshauer2012stable,camoriano2016nytro} and faster convergence~\cite{cutajar2016preconditioning, avron2016faster}. 
In particular, we note~\cite{fasshauer2012stable} which suggests approximations using top eigenvectors of the kernel matrix as a preconditioner, an idea closely related to EigenPro.

\section{Experimental Results}
\label{sec:expr}
%To underpin the adoption of EigenPro, this section analyzes the eigenspectrum of several real-world data sets. The analysis suggests that EigenPro preconditioning can increase the step size by a factor as large as three order of magnitude. We then investigate the overhead of the EigenPro methods, which turns out to be insignificant. Finally, we compare several scalable kernel methods with and without EigenPro acceleration. EigenPro preconditioned methods run substantially faster in all cases.
%; (4) the variance of the stochastic gradient with and without EigenPro is checked in Section \ref{sec:expr-vr}.
In this section, we will present a number of experimental results to evaluate EigenPro iteration on a range of datasets. 
% Specifically, 
% %could our approach search a much larger parameter space than the original kernel methods under a limited resource/time budget? The results are highly supportive. 
% the larger computational reach of our approach is implied by the substantially faster decrease of loss. Furthermore, running upon only a single GPU for a few hours, our methods (marginally) beat the state-of-the-art results of several scalable kernel methods using supercomputers or clusters of machines. Such result suggests that expanding the computational reach at an affordable cost is likely the key to the success of shallow learning.

\begin{wraptable}{r}{0.45\linewidth}
%\begin{table}[thb!]
\vspace{-4mm}
%\caption{Datasets}
%\label{tbl:dataset}
%\begin{tabular}{|p{25mm}|c|c|p{12mm}||c|}
\centering
\begin{adjustbox}{center}
\resizebox{7cm}{!}{
\begin{tabular}{|l|c|c|c|}
\hline
Name & $n$ & $d$ & Label \\ \hline % & $\gamma^{\dagger}$ \\ \hline
CIFAR-10
& $5 \times 10^4$ & 1024 & \{0,...,9\} \\ \hline % & 2e-2 \\ \hline
MNIST
& $6 \times 10^4$ & 784 & \{0,...,9\} \\ \hline % & 2e-2 \\ \hline
SVHN
& $7 \times 10^4$ & 1024 & \{1,...,10\} \\ \hline % & 2e-2 \\ \hline
HINT-S
& $2 \times 10^5$ & 425 & $\{0,1\}^{64}$ \\ \hline % & 2e-3 \\ \hline
TIMIT
& $1.1 \times 10^6$ & 440 & \{0,...,143\} \\ \hline % & 3e-3 \\ \hline
SUSY
& $5 \times 10^6$ & 18 & $\{0,1\}$ \\ \hline % & 2e-2 \\ \hline
HINT-M
& $7 \times 10^6$ & 246 & $[0,1]^{64}$ \\ \hline % & 2e-3 \\ \hline
MNIST-8M
& $8 \times 10^6$ & 784 & \{0,...,9\} \\ \hline % & 2e-2 \\ \hline
\end{tabular}
}
\end{adjustbox}
%\end{table}
\vspace{-5mm}
\end{wraptable}
\noindent {\bf Computing Resource.} All experiments were run on a single workstation equipped with 128GB main memory, two Intel Xeon(R) E5-2620 processors, and one Nvidia GTX Titan X (Maxwell) GPU.

\noindent {\bf Datasets.}
The table on the right summarizes
the datasets used in experiments.
For image datasets (MNIST~\cite{lecun1998gradient}, CIFAR-10~\cite{krizhevsky2009learning}, and SVHN~\cite{netzer2011reading}), color images are first transformed to grayscale images. We then rescale the range of each feature to $[0, 1]$.  For other datasets (HINT-S, HINT-M~\cite{healy2013algorithm}, TIMIT~\cite{garofolo1993darpa}, SUSY~\cite{baldi2014searching}), we normalize each feature by z-score. In addition, all multiclass labels are mapped to multiple binary labels. \newtxtblock

\noindent {\bf Metrics.} For datasets with multiclass or binary labels, we measure the training result by classification error ({\bf c-error}), the percentage of predicted labels that are incorrect; for datasets with real valued labels, we adopt the mean squared error ({\bf mse}).
\newtxtblock

\noindent {\bf Kernel methods.} For smaller datasets exact solution of kernel regularized least squares ({\bf KRLS}) gives the error close to optimal for kernel methods with the specific kernel parameters. To handle large dataset, we adopt primal space method, { Pegasos}~\cite{shalev2011pegasos} using the square loss and stochastic gradient. For even larger dataset, we combine SGD and Random Fourier Features~\cite{rahimi2007random} ({\bf RF}, see Appendix~\ref{ap:rf}) as in \cite{dai2014scalable,tu2016large}.
The results of these two methods are presented as the baseline. Then we apply EigenPro to Pegasos
%({\bf P-EigenPro})
and RF
%({\bf R-EigenPro})
as described in Section~\ref{sec:linear}. In addition, we compare the state-of-the-art results of other kernel methods to that of EigenPro in this section. \newtxtblock

\noindent {\bf Hyperparameters.} For consistent comparison, all iterative methods use mini-batch of size $m = 256$.
%and Gaussian kernel.
EigenPro  preconditioner is constructed using the top $k = 160$ eigenvectors of a subsampled dataset of size $M = 4800$. For EigenPro iteration  with random features, we  set the damping factor $\tau = \frac{1}{4}$. For primal EigenPro $\tau = 1$. \newtxtblock

\begin{wrapfigure}{r}{0.5\textwidth}
  \centering
  \vspace{-5mm}
  \includegraphics[width=0.4\textwidth]{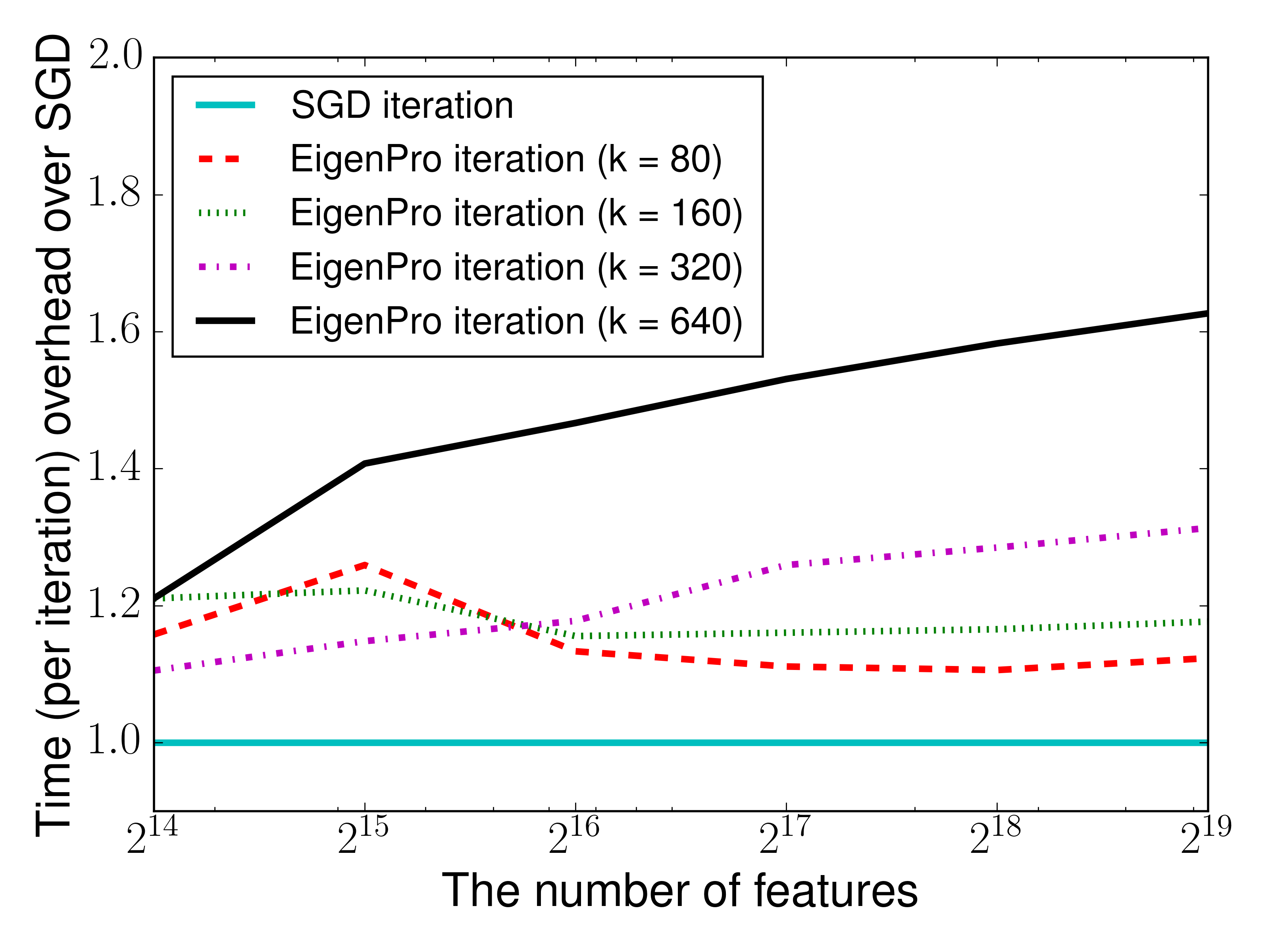}
  \vspace{-5mm}
\end{wrapfigure}
\noindent {\bf  Overhead of EigenPro iteration.}
The right side figure shows that the computational overhead of EigenPro iteration over the standard SGD ranged between $10\%$ and $50\%$. For $k = 160$ which is the default setting in all other experiments, EigenPro overhead is approximately $20\%$.
\newtxtblock
%%%%%%%%%%%%%% Overhead(E) %%%%%%%%%%%%%%%

%%%%%%%%%%%%%% Macro(S) %%%%%%%%%%%%%%%
\noindent {\bf Convergence acceleration by EigenPro for different kernels.}
Table~\ref{tbl:cmp-exact} presents the number of epochs needed  by EigenPro and Pegasos 
to reach the error of the optimal kernel classifier (computed by a direct method on these smaller datasets). The actual error can be found in Appendix~\ref{ap:kernel-select}.
%for reaching the optimal error.
%In comparison to Pegasos, EigenPro iteration saves 85\%-97\% epochs for settings using different kernels.
We see that EigenPro provides  acceleration of $6$ to $35$ times in terms of the number of epochs required without any loss of accuracy. 
The actual acceleration is about $20\%$ less due to the overhead of maintaining and applying a  preconditioner.
\begin{table*}[!ht]
%\vspace{-5mm}
\caption{Number of epochs to reach the optimal classification error (by KRLS)
}
\label{tbl:cmp-exact}
\begin{threeparttable}
\begin{adjustbox}{center}
\resizebox{13cm}{!}{
\begin{tabular}{|c|c||c|c||c|c||c|c|}
\hline
\multirow{2}{*}{Dataset} & \multirow{2}{*}{Size} & \multicolumn{2}{c||}{Gaussian Kernel} & \multicolumn{2}{c||}{Laplace Kernel} & \multicolumn{2}{c|}{Cauchy Kernel} \\ \cline{3-8} 
 &  & EigenPro & Pegasos & EigenPro & Pegasos & EigenPro & Pegasos \\ \hline
MNIST & $6 \times 10^4$ & \textbf{7} & 77 & \textbf{4} & 143 & \textbf{7} & 78 \\ \hline
CIFAR-10 & $5 \times 10^4$ & \textbf{5} & 56 & \textbf{13} & 136 & \textbf{6} & 107 \\ \hline
SVHN & $7 \times 10^4$ & \textbf{8} & 54 & \textbf{14} & 297 & \textbf{17} & 191 \\ \hline
HINT-S & $5 \times 10^4$ & \textbf{19} & 164 & \textbf{15} & 308 & \textbf{13} & 126 \\ \hline
\end{tabular}
}
\end{adjustbox}
% \begin{tablenotes}%[flushleft]
% \begin{minipage}{11cm}
% \footnotesize
% \item $*$ We use KRLS to obtain the optimal error. 
% \end{minipage}
% \end{tablenotes}
\end{threeparttable}
%\vspace{-2mm}
\end{table*}
%, see below.
%\newtxtblock
%%%%%%%%%%%%%% Macro(E) %%%%%%%%%%%%%%%

% %%%%%%%%%%%% Kernel Bandwidth (S) %%%%%%%%%%
\noindent {\bf Kernel bandwidth selection.}
We have investigated the impact of kernel bandwidth selection over convergence and performance for Gaussian kernel. As expected, kernel matrix with smaller bandwidth has slower eigenvalue decay, which in turn accelerates convergence of gradient descent.
However, selecting smaller bandwidth also decreases test set performance.
When the bandwidth is very small, the Gaussian classifier converges  to 1-nearest neighbor method, something which we observe in practice. While 1-NN classifier provides reasonable performance, it has up to twice the error of the optimal Bayes classifier in theory and far from 
 the carefully selected Gaussian kernel classifier in practice.
See Appendix~\ref{ap:bandwidth} for detailed results.
\newtxtblock
% %%%%%%%%%%%% Kernel Bandwidth (E) %%%%%%%%%%

%%%%%%%%%%%%%% Large(S) %%%%%%%%%%%%%%%
\noindent {\bf Comparisons on large datasets}.
On datasets involving up to a few million points, EigenPro consistently outperforms Pegasos/SGD-RF by a large margin when training with the same number of epochs (Table~\ref{tbl:err-epoch}). %Note with only $2\times 10^5$ random features, EigenPro-RF using full MINIST-8M and HINT-M yield significant lower errors than that of EigenPro using $1\times 10^6$ subsamples.
\begin{table*}[ht]
\caption{Error rate after 10 epochs / GPU hours (with Gaussian kernel)}
\label{tbl:err-epoch}
\begin{threeparttable}
\begin{adjustbox}{center}
\resizebox{15cm}{!}{
\begin{tabular}{|c|c|c||c|c|c|c||c|c|c|c|}
\hline
\multirow{2}{*}{Dataset} & \multirow{2}{*}{Size} & \multirow{2}{*}{Metric} & \multicolumn{2}{c|}{EigenPro} & \multicolumn{2}{c||}{Pegasos} & \multicolumn{2}{c|}{EigenPro-RF$^{\dagger}$} & \multicolumn{2}{c|}{SGD-RF$^{\dagger}$} \\ \cline{4-11} 
 &  &  & result & hours & result & hours & result & hours & result & hours \\ \hline
HINT-S & $2 \times 10^5$ & \multirow{4}{*}{c-error} & \textbf{10.0\%} & 0.1 & 11.7\% & 0.1 & 10.3\% & 0.2 & 11.5\% & 0.1 \\ \cline{1-2} \cline{4-11} 
TIMIT & $1 \times 10^6$ &  & \textbf{31.7\%} & 3.2 & 33.0\% & 2.2 & 32.6\% & 1.5 & 33.3\% & 1.0 \\ \cline{1-2} \cline{4-11}
\multirow{2}{*}{MNIST-8M} & $1 \times 10^6$ &  & \textbf{0.8\%} & 3.0 & 1.1\% & 2.7 & 0.8\% & 0.8 & 1.0\% & 0.7 \\ \cline{2-2} \cline{4-11} 
 & $8 \times 10^6$ &  & \multicolumn{2}{c|}{-} & \multicolumn{2}{c||}{-} & \textbf{0.7\%} & 7.2 & 0.8\% & 6.0 \\ \hline
\multirow{2}{*}{HINT-M} & $1 \times 10^6$ & \multirow{2}{*}{mse} & \textbf{2.3e-2} & 1.9 & 2.7e-2 & 1.5 & 2.4e-2 & 0.8 & 2.7e-2 & 0.6 \\ \cline{2-2} \cline{4-11} 
 & $7 \times 10^6$ &  & \multicolumn{2}{c|}{-} & \multicolumn{2}{c||}{-} & \textbf{2.1e-2} & 5.8 & 2.4e-2 & 4.1 \\ \hline
\end{tabular}
}
\end{adjustbox}
\begin{tablenotes}[flushleft]
\begin{minipage}{15cm}
\footnotesize
\item $\dagger$ We adopt $D = 2 \times 10^5$ random Fourier features.
\end{minipage}
\end{tablenotes}
\end{threeparttable}
%\vspace{-6mm}
\end{table*}

% When the resource/time budget is limited and the dataset is large,
% EigenPro can still substantially decrease the error rate at the cost of up to $26\%$ overhead (Table~\ref{tbl:err-epoch}). Note that for  MNIST-8M and HINT-M, GPU memory can no longer hold the entire dataset used as kernel centers, which makes running primal methods difficult. On the other hand random feature methods can still be used. Moreover, with only $2\times 10^5$ random features, they  outperform the primal method trained with $1 \times 10^6$ centers by a significant margin.
%%%%%%%%%%%%%% Large(E) %%%%%%%%%%%%%%%

%%%%%%%%%%%%%% Best(S) %%%%%%%%%%%%%%%
\noindent {\bf Comparisons to the state-of-the-art}. \label{sec:expr-sota}
In Table~\ref{tbl:best-result} we provide a comparison to  state-of-the-art results for large datasets recently reported in the kernel literature. All of them use significant computational resources and sometimes complex training procedures. We see that EigenPro  improves or matches performance performance on each dataset typically at a small fraction of the computational budget.
We notice that the very recent work~\cite{may2017kernel} achieves a better 30.9\% error rate on TIMIT (using an AWS cluster). It is not directly comparable to our result as it employs kernel features generated using a new supervised  feature selection method.  EigenPro can plausibly further improve the training error or decrease computational requirements using this new feature set.
\begin{table*}[!ht]
\caption{Comparison to large scale kernel results (Gaussian kernel)}
\label{tbl:best-result}
\begin{threeparttable}
\begin{adjustbox}{center}
\resizebox{15cm}{!}{
\begin{tabular}{|c|c||c|c|c||c|c|c|}
\hline
\multirow{2}{*}{Dataset} & \multirow{2}{*}{Size} & \multicolumn{3}{c||}{EigenPro (use 1 GTX Titan X)} & \multicolumn{3}{c|}{Reported  results} \\ \cline{3-8} 
 &  & error & GPU hours & epochs & source & error & description \\ \hline
\multirow{2}{*}{MNIST} & $1 \times 10^6$ & \textbf{0.70\%} & 4.8 & 16 & PCG~\cite{avron2016faster} & 0.72\% & \begin{tabular}[c]{@{}c@{}}1.1 hours (189 epochs)\\ on 1344 AWS vCPUs\end{tabular} \\ \cline{2-8} 
 & $6.7 \times 10^6$ & \textbf{0.80\%}$^\dagger$ & 0.8 & 10 & \cite{lu2014scale} & 0.85\% & \begin{tabular}[c]{@{}c@{}}less than 37.5 hours\\ on 1 Tesla K20m\end{tabular} \\ \hline
\multirow{2}{*}{TIMIT} & \multirow{2}{*}{$2 \times 10^6$} & \multirow{2}{*}{\begin{tabular}[c]{@{}c@{}}\textbf{31.7\%}\\ (32.5\%)$^\ddagger$\end{tabular}} & \multirow{2}{*}{3.2} & \multirow{2}{*}{10} & Ensemble~\cite{huang2014kernel} & 33.5\% & \begin{tabular}[c]{@{}c@{}}512 IBM\\ BlueGene/Q cores\end{tabular} \\ \cline{6-8} 
 &  &  &  &  & BCD~\cite{tu2016large} & 33.5\% & \begin{tabular}[c]{@{}c@{}}7.5 hours on \\ 1024 AWS vCPUs\end{tabular} \\ \hline
SUSY & $4 \times 10^6$ & \textbf{19.8\%} & 0.1 & 0.6 & Hierarchical~\cite{chen2016hierarchically} & $\approx 20\%$ & \begin{tabular}[c]{@{}c@{}}0.6 hours on \\ IBM POWER8\end{tabular} \\ \hline
\end{tabular}
}
\end{adjustbox}
\begin{tablenotes}[flushleft]
\begin{minipage}{15cm}
\footnotesize
\item $\dagger$ This result is produced by EigenPro-RF using $1\times 10^6$ data points.
\item $\ddagger$ Our TIMIT training set ($1\times 10^6$ data points) was generated following a standard practice in the speech community~\cite{povey2011kaldi} by
taking 10ms frames and dropping the glottal stop 'q' labeled frames in core test set ($1.2\%$ of total test set).
\cite{huang2014kernel} adopts 5ms frames, resulting in $2\times 10^6$ data points, and keeping the glottal stop 'q'.
Taking the worst case scenario for our setting, if we mislabel all glottal stops, the corresponding frame-level error will increase from $31.7\%$ to $32.5\%$.
\end{minipage}
\end{tablenotes}
\end{threeparttable}
%\vspace{-6mm}
\end{table*}
%%%%%%%%%%%%%% Best(E) %%%%%%%%%%%%%%%

\section{Conclusion and perspective}
In this paper we have considered a subtle trade-off between smoothness and computation for gradient-descent based method. While smooth output functions, such as those produced by kernel algorithms with smooth kernels,  are often desirable and help generalization (as, for example, encoded in the notion of algorithmic stability~\cite{bousquet2002stability}) there appears to be  a hidden  but very significant computational cost when the smoothness of the kernel  is mismatched with that of the target function. We argue and provide experimental evidence that these mismatches are common in standard classification problems. In particular, we view effectiveness of EigenPro as another piece of supporting evidence.

An important direction of future work is to understand whether this is a universal phenomenon encompassing a range of learning methods  or something pertaining to the kernel setting.  Specifically, the implications of this idea for deep neural networks need to be explored.  Indeed, there is a body of evidence indicating that  training neural networks results in highly non-smooth functions. For one thing, they can easily fit data even when the labels are randomly assigned~\cite{zhang2016understanding}.  Moreover, the pervasiveness of adversarial~\cite{szegedy2013intriguing} and even universal adversarial examples common to different networks~\cite{moosavi2016universal}  suggests that there are many directions non-smoothness in the neighborhood of nearly any data point. Why neural networks show generalization despite this evident non-smoothness, remains a key open question.

Finally, we have seen that training of kernel methods on large data can be significantly improved by  simple algorithmic modifications of first order iterative algorithms using limited second order information.  It appears that purely second order methods cannot provide major improvements  as low-rank approximations needed for dealing large data  discard information corresponding to the higher frequency components present in the data.
Better understanding of the computational and statistical issues and the trade-offs inherent in training,  would no doubt result in even better  shallow algorithms making them more competitive with deep networks on a given computational budget.

\section*{Acknowledgements}
We thank Adam Stiff and Eric Fosler-Lussier for preprocessing and providing the TIMIT dataset. We are also grateful to Jitong Chen and Deliang Wang for providing the HINT dataset. We thank the National Science Foundation for financial support (IIS 1550757 and CCF 	1422830) Part of this work was completed while the second author visited the Simons Institute at Berkeley.

\clearpage
\bibliographystyle{alpha2}
\bibliography{ref}
%\printbibliography

\clearpage
\begin{appendices}
\section{Scalable truncated SVD}
\label{ap:rsvd}
\begin{table*}[!ht]
%\begin{subalgorithm}{.5\textwidth}
%% RSVD
% \captionsetup{font=normalsize,singlelinecheck=off}
% \caption{{\bf Algorithm:} $\mathrm{RSVD}(X, k, M)$, adapted from \cite{halko2011finding}}\label{alg:rsvd}
%\begin{minipage}[t]{.5\linewidth}
\begin{minipage}[t]{.5\textwidth}
\begin{subalgorithm}{.95\linewidth}
\begin{algorithmic}[1]
  \ALG $\mathrm{RSVD}(X, k, M)$, adapted from \cite{halko2011finding}
  \INPUT
    $n \times d$ matrix $X$,
    number of eigen-directions $k$,
    subsample size $M$
  \OUTPUT
  	$(\be_1(H_M),\ldots,\be_k(H_M))$,
    $\mathrm{diag}(\lambda_1(H_M), \ldots, \lambda_k(H_M)), \lambda_{k+1}(H_M)$
    \STATE $X_M \leftarrow$ $M$ rows sampled from $X$ without replacement
    \STATE
    $[\lambda_1(X_M), \ldots, \lambda_{k+1}(X_M)]$, \\
    $[\bv_1(X_M),\ldots,\bv_{k}(X_M)]$ \\
    $\leftarrow \mathrm{rsvd}(X_M, k+1)$,
    non-Hermitian version of Algorithm 5.6 in \cite{halko2011finding} ($\bv_i(X) \defeq $ $i$-th right singular vector)
    \STATE $\lambda_i(H_M) \leftarrow \sqrt{\frac{n}{M}} \lambda_i(X_M)$, \\
    $\be_i(H_M) \leftarrow \bv_i(X_M)$ for $i = 1, \ldots, k$
\end{algorithmic}
\end{subalgorithm}
\end{minipage}
%\end{subalgorithm}%
%% NSVD
%\begin{subalgorithm}{.5\textwidth}
% \captionsetup{font=normalsize,singlelinecheck=off}
% \caption{{\bf Algorithm:} $\mathrm{NSVD}(X, k, M)$, adapted from \cite{williams2001using}}\label{alg:nsvd}
%\begin{minipage}[t]{0.95\linewidth}
\begin{minipage}[t]{.5\textwidth}
\begin{subalgorithm}{.95\linewidth}
\begin{algorithmic}[1]
  \ALG $\mathrm{NSVD}(X, k, M)$, adapted from \cite{williams2001using}
  \INPUT
    $n \times d$ matrix $X$,
    number of eigen-directions $k$,
    subsample size $M$
  \OUTPUT
  	$(\be_1(H_M),\ldots,\be_k(H_M))$,
    $\mathrm{diag}(\lambda_1(H_M), \ldots, \lambda_k(H_M)), \lambda_{k+1}(H_M)$
 
    \STATE $X_M \leftarrow$ $M$ rows sampled from $X$ without replacement
    \STATE $W \leftarrow \frac{n}{M} X_M X_M^T$
    \STATE
    $[\lambda_1(W), \ldots, \lambda_{k+1}(W)]$, \\
    $[\bv_1(W),\ldots,\bv_{k}(W)]$
    $\leftarrow \mathrm{svd}(W, k + 1)$
    \STATE $\lambda_i(H_M) \leftarrow \lambda_i(W)$, \\
    $\tilde{\be}_i \leftarrow \frac{n}{M} X_M^T \bv_i(W)$ for $i = 1, \ldots, k$
    \STATE $(\be_1(H_M), \ldots, \be_k(H_M)) \leftarrow$ \\
    $\mathrm{orthogonalize}(\tilde{\be}_1, \ldots, \tilde{\be}_k)$
\end{algorithmic}
\end{subalgorithm}
\end{minipage}
%\end{subalgorithm}
% \captionsetup{labelformat=alglabel}
% \caption{SVD Algorithms}%
% \label{alg:rsvd}%
\end{table*}

Given data matrix $X$, Algorithm RSVD computes its approximate top eigensystem using corresponding subsample covariance matrix $H_M \defeq \frac{1}{M} X_M^T X_M$. 
The algorithm has time complexity $O(Md\log{k} + (M + d) k^2)$ according to \cite{halko2011finding}.

The selection of subsample size $M$ depends on the eigenvalue decay rate of the covariance. Specifically, we choose $M = 4800$, which is much smaller than the size of training sets in our experiments. Thus, when $k = 160$, RSVD only takes a few minuets computation to complete even that the data (or kernel matrix) dimension $d$ exceeds $10^6$.

When computing the approximate top-eigensystem of the covariance for the EigenPro preconditioner, {N}ystr{\"o}m method based SVD (NSVD) is an alternative method with theoretical time complexity $O(Mdk + M^3)$ worse than that of RSVD.
However, its GPU implementation often outrun the corresponding RSVD implementation (due to applying SVD upon an $M \times M$ matrix instead of an $M \times d$ matrix). In practice, NSVD is able to obtain the approximate top-160 eigensystem of a kernel matrix with dimension $7 \cdot 10^6$ in less than 2.5 minuets using a single GTX Titan X (Maxwell).

Note that the results returned by NSVD normally involve larger approximation error than that by RSVD (using the same $M$). While in our experiments, this increase of the approximation error has negligible impact on the final regression/classification performance, as well as the convergence rate.

%%%%%%%%%%%%%% Feature Map (S) %%%%%%%%%%%%%
\section{Kernel-related feature maps}
\label{ap:rf}
\noindent {\bf Random Fourier features (RFF)~\cite{rahimi2007random}.}
The random Fourier feature map $\mathbb{R^N}\to \mathbb{R^d}$ is defined by
% to a shift invariant kernel $\rmk(\bx, \bx') = \rmk(\bx - \bx')$ for $\bx, \bx' \in \mathbb{R}^d$,
%\begin{equation} \label{eq:rff}
$$
\phi_{\mathrm{rff}}(\bx) \defeq \sqrt{2d^{-1}}(\cos(\bomega_1^T\bx + b_1), \ldots, \cos(\bomega_d^T\bx + b_d))^T
$$
%\end{equation}
Here $\bomega_1, \dots, \bomega_d$ are  sampled from the distribution $\frac{1}{2\pi} \int{\exp(-j \bomega^T \delta)\rmk(\delta)}{d\Delta}$ and $b_1, \ldots, b_d$ from uniform distribution on $[0, 2\pi]$.
It can be shown that the inner product in the feature space approximates  a Gaussian kernel $\rmk(\cdot, \cdot)$, i.e.,
%\begin{equation} \label{eq:rff-kernel}
$\lim_{D \rightarrow \infty}{\phi_{\mathrm{rff}}(\bx)^T \phi_{\mathrm{rff}}(\by)} = \rmk(\bx, \by)$.
%\end{equation}
\newtxtblock

\noindent {\bf Radial basis function network (RBF) features.}%~\cite{queback,camoriano2016nytro}.} 
This setting involves feature map related to kernel $\rmk(\cdot, \cdot)$, defined as
%\begin{equation} \label{eq:rbf}
$$
\phi_{\mathrm{rbf}}(\bx) \defeq
(\rmk(\bx, \bz_1), \dots, \rmk(\bx, \bz_d))^T
$$
%\end{equation}
where $\{ z_i \}_{i=1}^{d}$%, $z_i \in \mathbb{R}^d$
are network centers. Note there are various strategies for selecting centers, e.g., randomly sampling from $X$~\cite{camoriano2016nytro} or by computing K-means centers leading to different interpretations in terms of data-dependent kernels~\cite{queback}.
%%%%%%%%%%%%%% Feature Map (E) %%%%%%%%%%%%%

\section{Concentration bounds}
\label{ap:concentration}
\noindent {\bf Spectral norm of subsample covariance matrix.}
With subsample size $m$, $H_m \defeq \frac{1}{m} X_m^T X_m$ is the subsample covariance matrix corresponding to covariance $H$ defined in Eq.~\ref{eq:linear-hessian}. To bound its approximation error, $\norm{H_m - H}$, we need the following concentration theorem,
%%%% 
\begin{theorem*}[Matrix Bernstein~\cite{tropp2015introduction}]\label{clm:bernstein}
Let $S_1, \ldots, S_m$ be independent random Hermitian matrices with dimension $d \times d$. Assume that each matrix satisfies
\begin{equation} %\label{eq:bernstein-cond}
\mathbb{E}[S_i] = 0 ~~\mathit{and}
~~\norm{S_i} \leq L
~~\mathit{for}~ i = 1, \ldots, m 
\end{equation}
Then for any $t > 0$, random matrix $Z \defeq \sum_{i=1}^m{S_i}$ has bound
\begin{equation} %\label{eq:bernstein-bound}
P\{ \norm{Z} \geq t \}
\leq 2d \cdot \exp{\left( \frac{-t^2/2}{v(Z) + L t /3}\right)}
\end{equation}
where $v(Z) \defeq \norm{\mathbb{E}[Z^T Z]}$.
\end{theorem*}
%%%%
Applying Matrix Bernstein theorem to the subsample covariance $H_m$ leads to the following Lemma,
% (ToDo) Remove this claim if approximate preconditioner is not analyzed.
\begin{lemma}\label{clm:sample-cov}
If $\norm{\bx}^2_2 \leq \kappa$ for any $\bx\in X$ and $\lambda_1 = \norm{H}$, $\norm{H_m - H}$ has following upper bound with probability at least $1 - \delta$,
\begin{equation} \label{eq:sample-cov}
\begin{split}
\norm{H_m - H} 
& \leq \sqrt{\left (\frac{\lambda_1 + \kappa}{3m} \ln{\frac{2d}{\delta}} \right )^2 
+ \frac{2 \lambda_1 \kappa}{m}\ln{\frac{2d}{\delta}}}
+ \frac{\lambda_1 + \kappa}{3m} \ln{\frac{2d}{\delta}}
\end{split}
\end{equation}
\end{lemma}

\begin{proof}
Consider subsample covariance matrix $H_m = \frac{1}{m} \sum_{i=1}^m{\bx_i \bx_i^T}$ using $\bx_1, \ldots, \bx_m$ randomly sampled from dataset $X$.
For each sampled point $\bx_i, \ldots, \bx_m$, let
\begin{equation}
S_i \defeq \frac{1}{m} (\bx_i \bx_i^T - H)
\end{equation}
Clearly, $S_1, \ldots, S_m$ are independent random Hermitian matrices. Since $\mathbb{E}[\bx_i \bx_i^T] = \mathbb{E}[H_m] = H$ and $\norm{\bx_i}_2^2 \leq \kappa$ for $i = 1, \ldots, m$, we have
\begin{equation}
\mathbb{E}[S_i] = 0
~\mathit{and}~ \norm{S_i} \leq \frac{\lambda_1 + \kappa}{m}
~~for~ i = 1, \ldots, m
\end{equation}
%%%%%%%%%%%%
Furthermore, notice that for any $i = 1, \ldots, m$, we have
\begin{equation}
\begin{split}
\mathbb{E}[S_i^2] & = \frac{1}{m^2}
\mathbb{E}[\bx_i \bx_i^T \bx_i \bx_i^T
- \bx_i \bx_i^T H
- H \bx_i \bx_i^T
+ H^2] \\
& = \frac{1}{m^2} (\mathbb{E}[\norm{\bx_i}^2 \bx_i \bx_i^T]) - H^2) \\
& \preceq \frac{1}{m^2} (\kappa H - H^2)
\preceq \frac{\kappa}{m^2} H
\end{split}
\end{equation}
Hence the spectral norm has the following upper bound,
\begin{equation}
\norm{\mathbb{E}[S_i^2]} \leq \frac{\lambda_1 \kappa}{m^2}
\end{equation}
%%%%%%%%%%
The sum of the random matrices, $H_m - H = \sum_{i=1}^m{S_i}$, can be bounded as follows:
\begin{equation}
\begin{split}
v(H_m - H) &= \norm{\mathbb{E}[(H_m - H)^T (H_m - H)]} \\
& = \norm{\sum_{i=1}^m{\mathbb{E}[S_i^2]}}
\leq \sum_{i=1}^m{\norm{\mathbb{E}[S_i^2]}}
\leq \frac{\lambda_1 \kappa}{m}
\end{split}
\end{equation}
%%%%%%%%%%%
Now applying Matrix Bernstein Theorem yields
\begin{equation}\label{eq:hm-var-bound}
\begin{split}
P\{\norm{H_m - H} \geq t \}
\leq 2d \cdot 
\exp{\left( 
	\frac{-t^2/2}{v(H_m - H) + 		
    	\frac{(\lambda_1 + \kappa) t}{3m}} \right)}
\end{split}
\end{equation}
Let the right side of Eq.~\ref{eq:hm-var-bound} be $\delta$, we obtain that
\begin{equation}
\begin{split}
t & = \sqrt{\left (\frac{\lambda_1 + \kappa}{3m} \ln{\frac{2d}{\delta}} \right )^2 
+ \frac{2 \lambda_1 \kappa}{m}\ln{\frac{2d}{\delta}}}
+ \frac{\lambda_1 + \kappa}{3m} \ln{\frac{2d}{\delta}}
\end{split}
\end{equation}
Therefore, with probability at least $1 - \delta$, we have
\begin{equation}
\norm{H_m - H} \leq t
\end{equation}
%%%%%%%%%%%
\end{proof}

\noindent {\bf Spectral norm of subsample covariance operator.}
For EigenPro iteration in RKHS, its step size selection is related to covariance operator $\rmH$ defined in Eq.~\ref{eq:cov-op}. Similarly, under the SGD setting, we need to bound $\norm{\rmH_m - \rmH}$ where $\rmH_m f(x) \defeq \frac{1}{m} \sum_{i=1}^m{k(x, x_i)f(x_i)}$ is the corresponding subsample covariance operator. Thus we introduce the following Bernstein inequality for random operators.
%Its key idea is to replace the matrix dimension $d$ in Matrix Bernstein theorem with intrinsic dimension of matrix/operator defined in~\cite{tropp2015introduction}.
%According to \cite{rosasco2010learning}, kernel integral operator $\rmH$ has bounded trace and spectral norm. Especially for Gaussian kernel, $tr(\rmH^2) \leq tr(\rmH) = 1$. Thus $\rmd \leq \norm{\rmH}^{-2}$.
%In RKHS kernel setting, the intrinsic dimension is usually well bounded. Hence it allows us to investigate the covariance operator defined on RKHS space of infinite dimension.
This inequality differs from its vector space version (Matrix Bernstein) by replacing the space dimension with the intrinsic dimension, defined as
\begin{equation} \label{eq:intdim}
\rmd(V) \defeq \frac{tr(\rmV)}{\norm{\rmV}}
\end{equation}

\begin{theorem*}[Operator Bernstein, adapted from~\cite{minsker2017some,tropp2015introduction}]\label{clm:op-bernstein}
Let $S_i: \mathcal{H} \rightarrow \mathcal{H}, i = 1, \ldots, m$ be independent random Hermitian operators. Assume that each operator satisfies
\begin{equation} \label{eq:op-bernstein-cond}
\mathbb{E}[S_i] = 0 ~~\mathit{and}
~~\norm{S_i} \leq L
~~\mathit{for}~ i = 1, \ldots, m 
\end{equation}
Consider random Hermitian operator $\rmZ \defeq \sum_{i=1}^m{S_i}$. Its variance $V \defeq \mathbb{E}[\rmZ^2]$ is also an operator on $\mathcal{H}$. For convenience, let the intrinsic dimension (Eq.~\ref{eq:intdim}) and the spectral norm of the variance operator be
$$
\rmd \defeq \rmd(V)
~~\mathit{and}~~
v \defeq \norm{V}
$$
Then for any $t \geq \sqrt{v} + L/3$, we have
\begin{equation} \label{eq:op-bernstein-bound}
P\{ \norm{\rmZ} \geq t \}
\leq 8\rmd \cdot \exp{\left( \frac{-t^2/2}{v + L t /3}\right)}
\end{equation}
\end{theorem*}

This theorem can be applied to bound $\norm{\rmH_m - \rmH}$,
\begin{lemma}\label{clm:sample-op-cov}
If $k(\bx, \bx) \leq \kappa$ for any $\bx\in X$ and $\lambda_1 = \norm{\rmH}$, $\norm{\rmH_m - \rmH}$ has following upper bound with probability at least $1 - \delta$,
\begin{equation} \label{eq:sample-op-cov}
\begin{split}
\norm{\rmH_m - \rmH} 
& \leq \sqrt{\left (\frac{\lambda_1 + \kappa}{3m} \ln{\frac{8 \rmd}{\delta}} \right )^2 
+ \frac{2 \lambda_1 \kappa}{m}\ln{\frac{8 \rmd}{\delta}}}
+ \frac{\lambda_1 + \kappa}{3m} \ln{\frac{8 \rmd}{\delta}}
\end{split}
\end{equation}
%for any $\delta \leq 8\rmd \cdot e^2$.
\end{lemma}

\begin{proof}
Consider subsample covariance operator $\rmH_m f(\bx) \defeq \frac{1}{m} \sum_{i=1}^m{k(\bx, \bx_i)f(\bx_i)}$ defined by $\bx_1, \ldots, \bx_m$ randomly sampled from dataset $X$.
For each sampled point $\bx_i, \ldots, \bx_m$, let
\begin{equation}
\rmS_i f(\bx) \defeq \frac{1}{m} (k(\bx, \bx_i)f(\bx_i) - \rmH f(\bx))
\end{equation}
Clearly, $\rmS_1, \ldots, \rmS_m$ are independent random Hermitian operators. Since $\mathbb{E}[k(\bx, \bx_i)f(\bx_i)] = \rmH f(\bx)$ and $\norm{k(x_i, \cdot)}_\mathcal{H} = k(\bx_i, \bx_i) \leq \kappa$ for $i = 1, \ldots, m$, we have
\begin{equation}
\mathbb{E}[\rmS_i] = 0
~\mathit{and}~ \norm{\rmS_i} \leq \frac{\lambda_1 + \kappa}{m}
~~for~ i = 1, \ldots, m
\end{equation}
%%%%%%%%%%%%
Furthermore, for any $f \in \mathcal{H}$, the variance operator of $\rmS_i$ equals
\begin{equation}
\begin{split}
\mathbb{E}[\rmS_i^2] f(\bx) &
= \frac{1}{m^2} (\mathbb{E}[k(\bx_i, \bx_i) k(\bx, \bx_i)f(\bx_i)] - \rmH^2 f(\bx)) \\
& = \frac{1}{m^2} (k(\bx_i, \bx_i) \rmH f(\bx) - \rmH^2 f(\bx))
\end{split}
\end{equation}
As $k(\bx_i, \bx_i) \leq \kappa$ and $\rmH$ is positive semi-definite, this variance operator is bounded by
$$
\mathbb{E}[\rmS_i^2]
\preceq \frac{1}{m^2} (\kappa \rmH - \rmH^2)
\preceq \frac{\kappa}{m^2} \rmH
$$
Hence its spectral norm has upper bound,
\begin{equation}
\norm{\mathbb{E}[\rmS_i^2]} \leq \frac{\lambda_1 \kappa}{m^2}
\end{equation}
%%%%%%%%%%
Since $\rmS_i$ and $\rmS_j$ are independent for $i \neq j$, $\mathbb{E}[\rmS_i \rmS_j] f(x) \equiv 0$ almost everywhere. Then the sum of the random operators, $\rmH_m - \rmH = \sum_{i=1}^m{\rmS_i}$, can be bounded as follows:
\begin{equation}
\begin{split}
v(\rmH_m - \rmH) &= \norm{\sum_{i=1}^m{\mathbb{E}[\rmS_i^2]}}
\leq \sum_{i=1}^m{\norm{\mathbb{E}[\rmS_i^2]}}
\leq \frac{\lambda_1 \kappa}{m}
\end{split}
\end{equation}
%%%%%%%%%%%
Now we can apply the Operator Bernstein on $\rmH_m - \rmH$ which yields concentration bound,
\begin{equation}\label{eq:op-hm-var-bound}
\begin{split}
P\{\norm{\rmH_m - \rmH} \geq t \}
\leq 8 \rmd \cdot 
\exp{\left( 
	\frac{-t^2/2}{v(\rmH_m - \rmH) + 		
    	\frac{(\lambda_1 + \kappa) t}{3m}} \right)}
\end{split}
\end{equation}
for any $t \geq \sqrt{v(\rmH_m - \rmH)} + \frac{(\lambda_1 + \kappa)}{3m}$.
Let the right side of Eq.~\ref{eq:op-hm-var-bound} be $\delta$, we obtain that
\begin{equation}
\begin{split}
t & = \sqrt{\left (\frac{\lambda_1 + \kappa}{3m} \ln{\frac{8 \rmd}{\delta}} \right )^2 
+ \frac{2 \lambda_1 \kappa}{m}\ln{\frac{8 \rmd}{\delta}}}
+ \frac{\lambda_1 + \kappa}{3m} \ln{\frac{8 \rmd}{\delta}}
\end{split}
\end{equation}
Therefore, for any $\delta \leq 8\rmd \cdot e$, with probability at least $1 - \delta$, we have
\begin{equation}
\norm{\rmH_m - \rmH} \leq t
\end{equation}
According to the definition of the intrinsic dimension, $\rmd \geq 1$. Thus $\delta \leq 8\rmd \cdot e$ is true for any $\delta \in [0, 1]$.
\end{proof}

\noindent {\bf Upper bound on the spectral norm of subsample covariance.}
%Combining Lemma~\ref{clm:sample-cov} 
Applying Lemma~\ref{clm:sample-cov} to $\norm{H_m - H}$ and using the inequality $\norm{H_m} \leq \norm{H} + \norm{H_m - H}$,
%and inequality $\sqrt{a^2 + b^2} \leq |a| + |b|$ 
one can obtain the following
\begin{lemma}\label{clm:bernstein-sgd}
If $\norm{\bx}^2_2 \leq \kappa$ for any $\bx\in X$, then with probability at least $1 - \delta$,
\begin{equation} \label{eq:bernstein-sgd}
\begin{split}
\norm{H_m}  \leq \lambda_1 + \frac{2 (\lambda_1 + \kappa)}{3m} \ln{\frac{2d}{\delta}}  
 + \sqrt{\frac{2 \lambda_{1} \kappa}{m} \ln{\frac{2d}{\delta}}}
\end{split}
\end{equation}
\end{lemma}

Similarly, we can bound the spectral norm of the subsample covariance operator $\rmH_m$ by using Lemma~\ref{clm:sample-op-cov}.
\begin{lemma}\label{clm:bernstein-sgd-op}
If $k(\bx, \bx) \leq \kappa$ for any $\bx\in X$, then with probability at least $1 - \delta$,
\begin{equation} %\label{eq:bernstein-sgd}
\begin{split}
\norm{\rmH_m}  \leq \lambda_1 + \frac{2 (\lambda_1 + \kappa)}{3m} \ln{\frac{8 \rmd}{\delta}}  
 + \sqrt{\frac{2 \lambda_{1} \kappa}{m} \ln{\frac{8 \rmd}{\delta}}}
\end{split}
\end{equation}
\end{lemma}

\section{\texorpdfstring{$\norm{\balpha}_{2}$} \ \  during the training of (primal) kernel method}
\label{ap:weight}

\begin{figure}[!hp]
  \centering
  \begin{minipage}[b]{0.45\textwidth}
    \includegraphics[width=\textwidth]{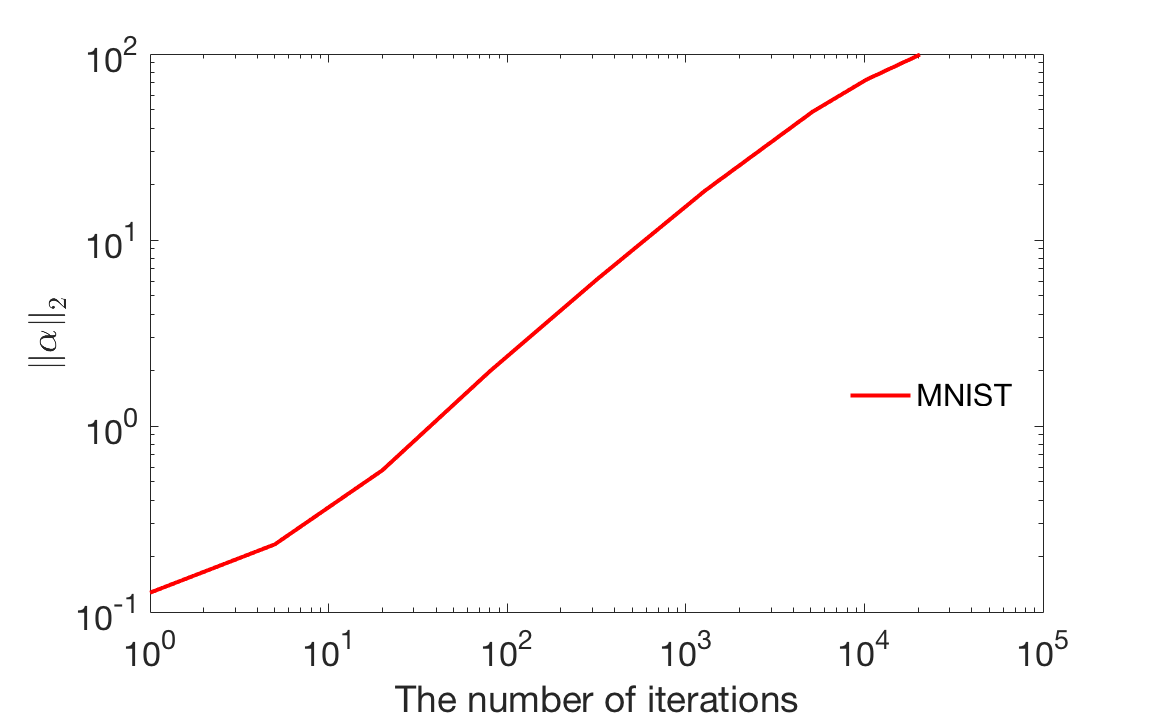}
  \end{minipage}
  \hfill
  \begin{minipage}[b]{0.45\textwidth}
    \includegraphics[width=\textwidth]{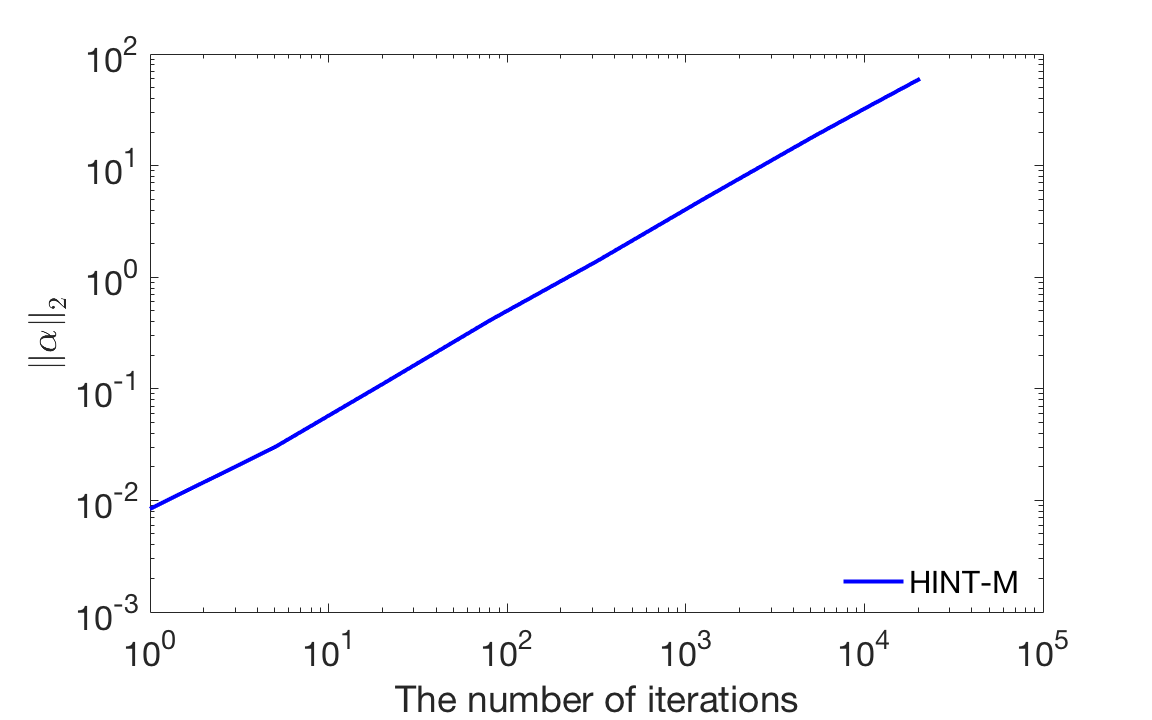}
  \end{minipage}
  \caption{Change of $\norm{\alpha}_2$ during graident descent training (Pegasos) on subsample datasets (10000 points)}
\end{figure}
Here $\balpha = (\alpha_1, \ldots, \alpha_n)^T$ is the representation of the solution $f$ under basis $\{ k(\bx_i, \cdot) \}_{i=1}^n$ such that $f = \sum_{i=1}^n {\alpha_i k(\bx_i, \cdot)}$. Therefore, $\norm{\balpha}_2 = \norm{f}_{L^2}$.
%%%%%%%%%%%%%% Weight norm (E) %%%%%%%%%%%%%

%%%%%%%%%%%%%% Kernel (S) %%%%%%%%%%%%%%%
\section{Kernel selection}
\label{ap:kernel-select}
%\noindent {\bf Kernel selection.}
\begin{wraptable}{r}{0.5\linewidth}
%\vspace{-2mm}
\centering
\caption*{Best classification error (for KRLS)}
\resizebox{\linewidth}{!}{%
\begin{tabular}{|c|c||c|c|c|}
\hline
Dataset & Size & Gaussian & Laplace & Cauchy \\ \hline
MNIST & $6 \times 10^4$ & \textbf{1.3\%} & 1.6\% & 1.5\% \\ \hline
CIFAR-10 & $5 \times 10^4$ & 49.5\% & 48.9\% & \textbf{48.4\%} \\ \hline
SVHN & $7 \times 10^4$ & \textbf{18.0\%} & 18.5\% & \textbf{18.0\%} \\ \hline
HINT-S & $5 \times 10^4$ & \textbf{11.3\%} & 11.4\% & \textbf{11.3\%} \\ \hline
\end{tabular}
}%
\vspace{-2mm}
\end{wraptable}
The table on the right side compares optimal performance for three different kernels: the Gaussian kernel $k_1(x, y) \defeq \exp(-\frac{\norm{x - y}^2}{2\sigma^2})$, the Laplace kernel $k_2(x, y) \defeq \exp(-\frac{\norm{x - y}}{\sigma})$, and the Cauchy kernel $k_3(x, y) \defeq (1 + \frac{\norm{x - y}^2}{\sigma^2})^{-1}$.
With bandwidth $\sigma$ selected by cross validation, the performance of the Gaussian kernel is generally comparable to that of the Cauchy kernel. The Laplace kernel performs worst possibly due to its non-smoothness. 
%Thus we will mainly use Gaussian kernel in the rest experiments.
%\newtxtblock
%%%%%%%%%%%%%% Kernel (E) %%%%%%%%%%%%%%%

%%%%%%%%% Comparing Kernel (S) %%%%%%%%
\begin{table}[!ht]
\centering
\caption{Classification error on MNIST with different kernels}
\label{tbl:kernel}
\begin{adjustbox}{center}
\resizebox{15cm}{!}{
\begin{tabular}{|c||c|c|c|c||c|c|c|c||c|c|c|c|}
\hline
\multirow{3}{*}{$N_{Epoch}$} & \multicolumn{4}{c||}{Gaussian Kernel ($\sigma^2 = 25$)} & \multicolumn{4}{c||}{Laplace Kernel ($\sigma = 10$)} & \multicolumn{4}{c|}{Cauchy Kernel ($\sigma^2 = 40$)} \\ \cline{2-13} 
 & \multicolumn{2}{c|}{EigenPro} & \multicolumn{2}{c||}{Pegasos} & \multicolumn{2}{c|}{EigenPro} & \multicolumn{2}{c||}{Pegasos} & \multicolumn{2}{c|}{EigenPro} & \multicolumn{2}{c|}{Pegasos} \\ \cline{2-13} 
 & train & test & train & test & train & test & train & test & train & test & train & test \\ \hline
1 & 0.92\% & 2.03\% & 5.12\% & 5.21\% & 0.16\% & 2.06\% & 7.39\% & 7.21\% & 0.41\% & 1.91\% & 6.05\% & 5.96\% \\ \hline
5 & 0.10\% & 1.44\% & 2.36\% & 2.84\% & 0.0\% & 1.62\% & 3.42\% & 4.26\% & 0.0\% & 1.31\% & 2.76\% & 3.44\% \\ \hline
10 & 0.01\% & 1.23\% & 1.58\% & 2.32\% & 0.0\% & 1.58\% & 2.18\% & 3.39\% & 0.0\% & 1.32\% & 1.76\% & 2.57\% \\ \hline
20 & 0.0\% & \textbf{1.20\%} & 0.90\% & 1.93\% & 0.0\% & \textbf{1.57\%} & 1.09\% & 2.57\% & 0.0\% & 1.30\% & 0.91\% & 2.12\% \\ \hline
40 & 0.0\% & 1.20\% & 0.39\% & 1.65\% & 0.0\% & 1.57\% & 0.32\% & 2.14\% & 0.0\% & 1.30\% & 0.30\% & 1.78\% \\ \hline
80 & 0.0\% & 1.23\% & 0.14\% & 1.41\% & 0.0\% & 1.57\% & 0.03\% & 1.85\% & 0.0\% & \textbf{1.29\%} & 0.06\% & 1.56\% \\ \hline
160 & 0.0\% & 1.21\% & 0.03\% & 1.24\% & 0.0\% & 1.57\% & 0.0\% & 1.71\% & 0.0\% & 1.29\% & 0.01\% & 1.36\% \\ \hline
\end{tabular}
}
\end{adjustbox}
\end{table}

%%%%%%%%%---------%%%%%%%%------%%%%%%%%%
\begin{table}[!ht]
\centering
\caption{L2 loss on MNIST with different kernels}
\label{tbl:kernel-l2}
\begin{adjustbox}{center}
\resizebox{15cm}{!}{
\begin{tabular}{|c||c|c|c|c||c|c|c|c||c|c|c|c|}
\hline
\multirow{3}{*}{$N_{Epoch}$} & \multicolumn{4}{c||}{Gaussian Kernel ($\sigma^2 = 25$)} & \multicolumn{4}{c||}{Laplace Kernel ($\sigma = 10$)} & \multicolumn{4}{c|}{Cauchy Kernel ($\sigma^2 = 40$)} \\ \cline{2-13} 
 & \multicolumn{2}{c|}{EigenPro} & \multicolumn{2}{c||}{Pegasos} & \multicolumn{2}{c|}{EigenPro} & \multicolumn{2}{c||}{Pegasos} & \multicolumn{2}{c|}{EigenPro} & \multicolumn{2}{c|}{Pegasos} \\ \cline{2-13} 
 & train & test & train & test & train & test & train & test & train & test & train & test \\ \hline
1 & 2.4e-2 & 3.2e-2 & 6.9e-2 & 6.8e-2 & 2.4e-2 & 3.5e-2 & 9.4e-2 & 9.3e-2 & 1.8e-2 & 3.0e-2 & 7.9e-2 & 7.8e-2 \\ \hline
5 & 8.6e-3 & 2.4e-2 & 4.0e-2 & 4.3e-2 & 1.7e-3 & 2.5e-2 & 5.3e-2 & 5.6e-2 & 3.0e-3 & 2.2e-2 & 4.4e-2 & 4.7e-2 \\ \hline
10 & 4.3e-3 & 2.2e-2 & 3.1e-2 & 3.6e-2 & 1.0e-4 & \textbf{2.4e-2} & 4.0e-2 & 4.6e-2 & 8.0e-4 & 2.1e-2 & 3.3e-2 & 3.9e-2 \\ \hline
20 & 1.8e-3 & \textbf{2.1e-2} & 2.3e-2 & 3.1e-2 & 0 & 2.4e-2 & 2.8e-2 & 3.9e-2 & 1.0e-4 & \textbf{2.0e-2} & 2.3e-2 & 3.2e-2 \\ \hline
40 & 6.1e-4 & 2.1e-2 & 1.6e-2 & 2.7e-2 & 0 & 2.4e-2 & 1.7e-2 & 3.2e-2 & 0 & 2.0e-2 & 1.5e-2 & 2.7e-2 \\ \hline
80 & 2.2e-4 & 2.1e-2 & 9.7e-3 & 2.4e-2 & 0 & 2.4e-2 & 8.3e-3 & 2.8e-2 & 0 & 2.0e-2 & 7.8e-3 & 2.4e-2 \\ \hline
160 & 8.0e-5 & 2.1e-2 & 5.1e-3 & 2.2e-2 & 0 & 2.4e-2 & 2.6e-3 & 2.6e-2 & 0 & 2.0e-2 & 3.1e-3 & 2.2e-2 \\ \hline
\end{tabular}
}
\end{adjustbox}
\end{table}

\section{Kernel selection on convergence and performance}
Table~\ref{tbl:kernel} and Table~\ref{tbl:kernel-l2} presents detailed results on MNIST with the kernel bandwidth $\sigma$ selected by cross validation. Results with all kernels generally show comparable convergence rate and performance. However, the performance of Gaussian kernel is better overall, a pattern we observed on other datasets as well.
%%%%%%%%% Comparing Kernel (E) %%%%%%%%

%%%%%%%%%%%%%% Eigenspectrum(S) %%%%%%%%%%%%%%%
\begin{figure}[!htpb]
    \centering
    
    \begin{subfigure}[t]{0.5\linewidth}
      \centering
       \includegraphics[width=0.8\linewidth, scale=1.0]{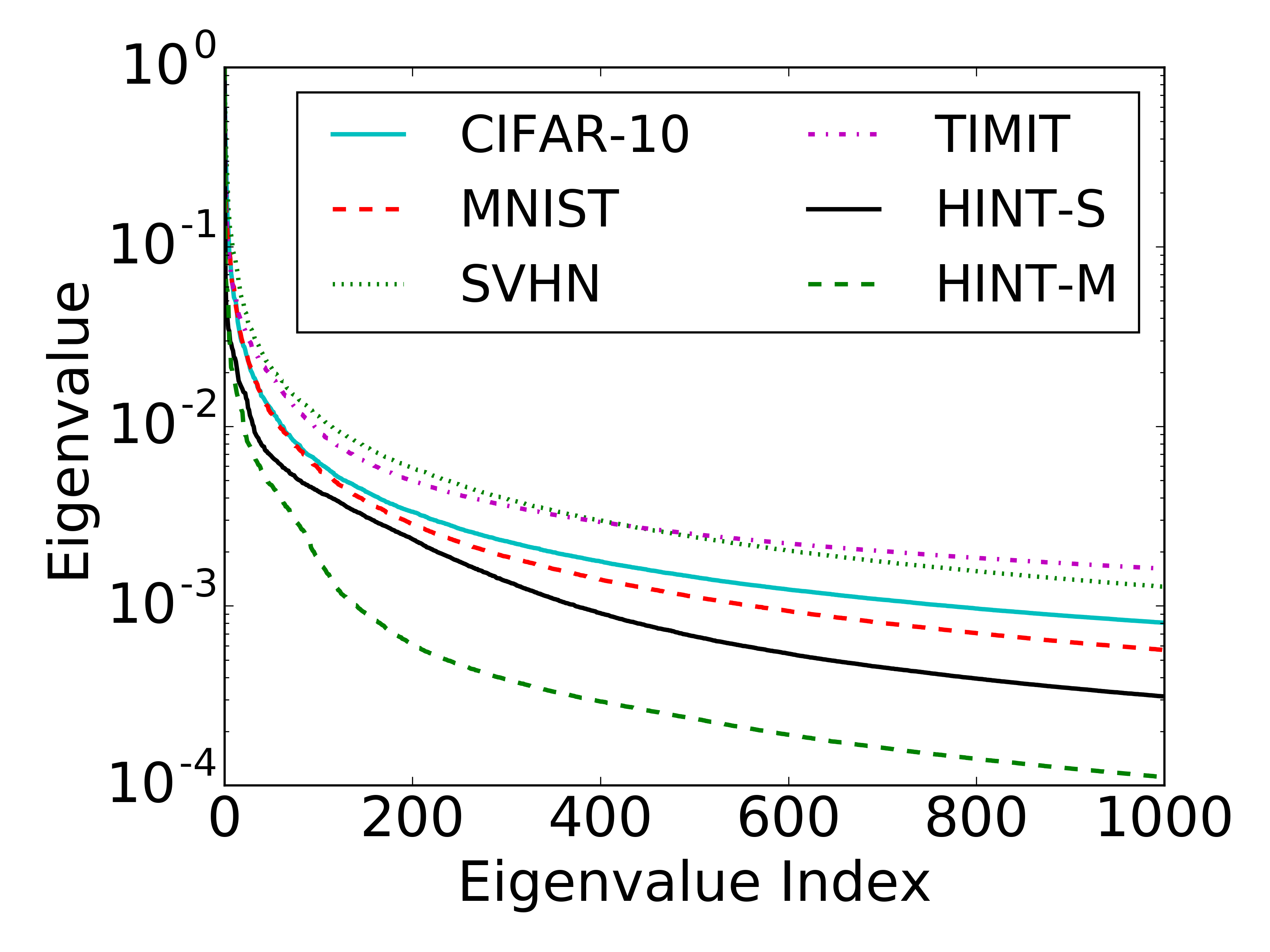}
       \vspace{-4mm}
        \caption{\small Different datasets, subsample size $N=2\cdot10^4$}
       \label{fig:es}
    \end{subfigure}%
    ~ 
    \begin{subfigure}[t]{0.5\linewidth}
      \centering
      \includegraphics[width=0.8\linewidth,scale=1.0]{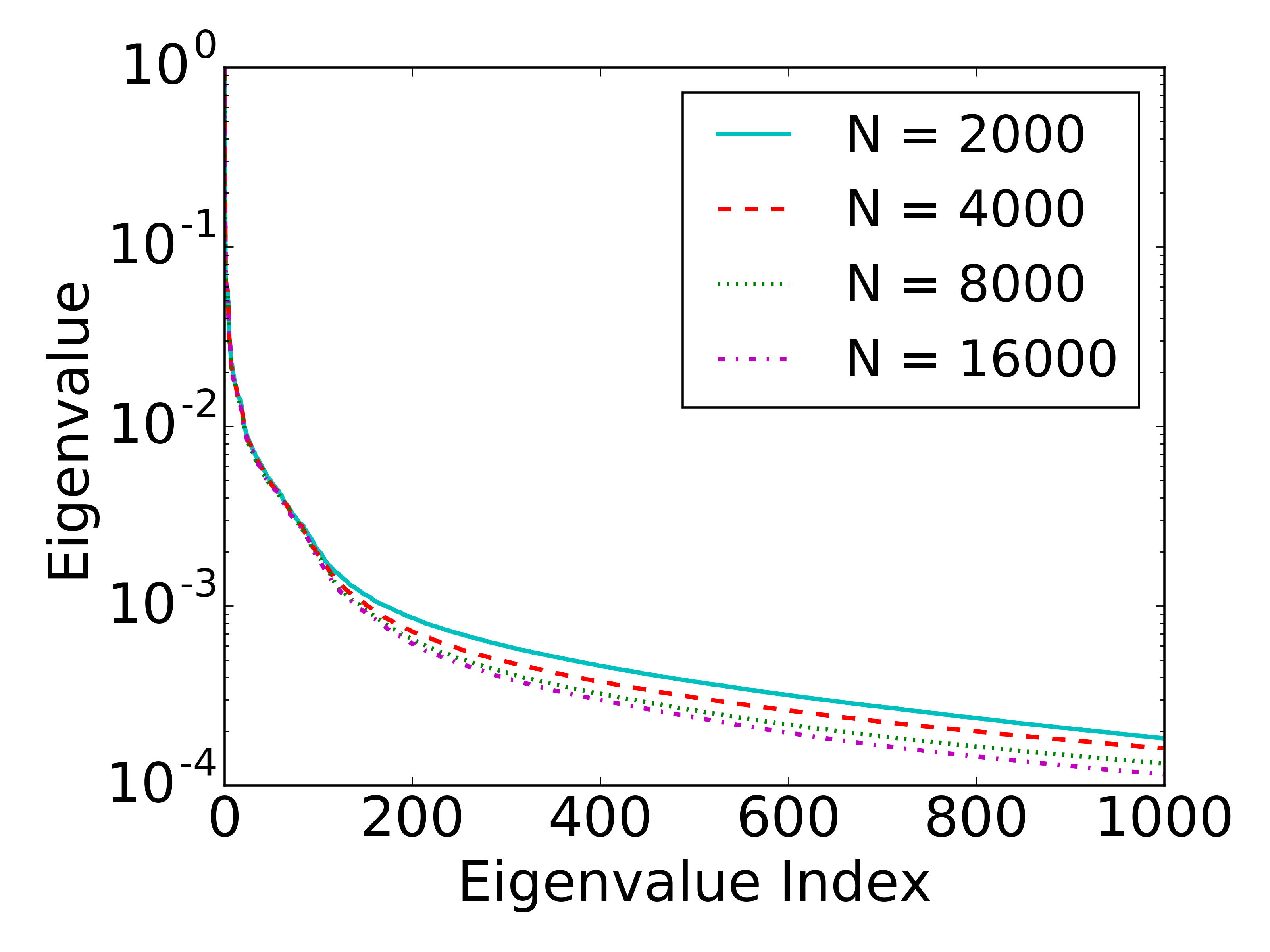}
      \vspace{-4mm}
      \caption{\small Same data set, subsets of different sizes ($N$) }
      \label{fig:es-hint-m}
    \end{subfigure}%
    \vspace{-4mm}
    \caption{Eigenspectrum of the kernel matrices}
    % \label{fig:eigen}
\end{figure}
\section{Eigenvalues of (Gaussian) kernel matrices}
%\noindent {\bf Eigenvalues of (Gaussian) kernel matrices.}
\begin{wraptable}{r}{0.6\linewidth}
\vspace{-5mm}
\centering
\caption*{Eigenvalue ratio of kernel matrices}
\label{tbl:es-ratio}
\resizebox{\linewidth}{!}{%
\begin{tabular}{|c|c|c|c|c|c|}
\hline
Ratio & CIFAR & MNIST & SVHN & TIMIT & HINT-S \\ \hline
$\lambda_1 / \lambda_{41}$ & 68 & 69 & 39 & 45 & 128 \\ \hline
$\lambda_1 / \lambda_{81}$ & 128 & 135 & 72 & 84 & 200 \\ \hline
$\lambda_1 / \lambda_{161}$ & 245 & 280 & 138 & 166 & 338 \\ \hline
$\lambda_1 / \lambda_{321}$ & 464 & 567 & 270 & 290 & 800 \\ \hline
\end{tabular}
}%
\vspace{-4mm}
\end{wraptable}
For each dataset, the eigenspectrum of the corresponding kernel matrix directly determines the impact of EigenPro on convergence. 
%Thus we calculate a kernel matrix for each dataset by 4000 subsamples from it. 
Figure \ref{fig:es} shows the normalized eigenspectra ($\lambda_i / \lambda_1$) of these kernel matrices. We see that the spectrum of each dataset drops sharply for the $200$ eigenvalues, making EigenPro highly effective.
Table above lists the eigenvalue ratios corresponding to different $k$. For example, with $k=160$, EigenPro for RF can in theory increase the step accelerate training by a factor of $338$ for training on HINT-S. Actual acceleration is not as large but still significant. Note that given small subsample size we expect that lower eigenvalues are probably not reflective of the full large kernel matrix. Still, even lower eigenvalues match closely when computed from subsamples of different size (Figure~\ref{fig:es-hint-m}).
%%%%%%%%%%%%%% Eigenspectrum(E) %%%%%%%%%%%%%%%

%%%%%%%%%%%% Kernel Bandwidth (S) %%%%%%%%%%
\section{Kernel bandwidth selection}
\label{ap:bandwidth}
%\noindent {\bf  Kernel bandwidth selection.}
Here we investigate the impact of kernel bandwidth selection over convergence and performance. Table~\ref{tbl:bandwidth} compares Pegasos and EigenPro iteration using Gaussian kernel with three different bandwidths $\sigma^2$. When $\sigma^2 = 25$, EigenPro reaches best error rate 1.20\% on testing data after 20 epochs training.
%Note that one EigenPro iteration outperforms 10 Pegasos iteration on both classification error and L2 loss.
Using a smaller bandwidth $\sigma^2 = 5$, EigenPro reaches its optimal in only 5 epochs. But its error rate 1.83\% is significantly worse than that of $\sigma^2 = 25$.
%Also, smaller bandwidth makes EigenPro even more effective. Now one EigenPro iteration outperforms 40 Pegasos iterations by a large margin.
In sum, using smaller kernel bandwidth leads to slower eigenvalue decay, which in turn improves convergence of gradient-based methods. While selecting bandwidth for faster convergence does not guarantee better generalization performance (as shown in Table~\ref{tbl:bandwidth}).
%Similar result is observed on TIMIT dataset (Table~\ref{tbl:bandwidth-timit}).
\newtxtblock
\begin{table}[!ht]
\centering
\caption{Impact of kernel bandwidth on classification error and L2 loss (MNIST)}
\label{tbl:bandwidth}
\begin{adjustbox}{center}
\resizebox{13cm}{!}{
\begin{tabular}{|c|c||c|c|c|c||c|c|c|c|}
\hline
\multirow{3}{*}{$\sigma^2$} & \multirow{3}{*}{$N_{Epoch}$} & \multicolumn{4}{c||}{EigenPro} & \multicolumn{4}{c|}{Pegasos} \\ \cline{3-10} 
 &  & \multicolumn{2}{c|}{c-error} & \multicolumn{2}{c||}{L2 loss} & \multicolumn{2}{c|}{c-error} & \multicolumn{2}{c|}{L2 loss} \\ \cline{3-10} 
 &  & train & test & train & test & train & test & train & test \\ \hline
\multirow{5}{*}{50} & 1 & 1.58\% & 2.39\% & 3.4e-2 & 4.0e-2 & 7.29\% & 6.86\% & 9.0e-2 & 8.8e-2 \\ \cline{2-10} 
 & 5 & 0.41\% & 1.66\% & 1.7e-2 & 2.9e-2 & 3.96\% & 4.17\% & 5.7e-2 & 5.7e-2 \\ \cline{2-10} 
 & 10 & 0.15\% & 1.48\% & 1.1e-2 & 2.6e-2 & 2.92\% & 3.36\% & 4.7e-2 & 4.9e-2 \\ \cline{2-10} 
 & 20 & 0.06\% & 1.41\% & 6.6e-3 & 2.4e-2 & 2.12\% & 2.66\% & 3.8e-2 & 4.2e-2 \\ \cline{2-10} 
 & 40 & 0.01\% & \textbf{1.25\%} & 3.3e-3 & 2.4e-2 & 1.45\% & 2.18\% & 3.0e-2 & 3.6e-2 \\ \hline
\hline
\multirow{5}{*}{25} & 1 & 0.92\% & 2.03\% & 2.4e-2 & 3.2e-2 & 5.12\% & 5.21\% & 6.9e-2 & 6.8e-2 \\ \cline{2-10} 
 & 5 & 0.10\% & 1.44\% & 8.6e-3 & 2.4e-2 & 2.36\% & 2.84\% & 4.0e-2 & 4.3e-2 \\ \cline{2-10} 
 & 10 & 0.01\% & 1.23\% & 4.3e-3 & 2.2e-2 & 1.58\% & 2.32\% & 3.1e-2 & 3.6e-2 \\ \cline{2-10} 
 & 20 & 0.0\% & \textbf{1.20\%} & 1.8e-3 & 2.1e-2 & 0.90\% & 1.93\% & 2.3e-2 & 3.1e-2 \\ \cline{2-10} 
 & 40 & 0.0\% & 1.20\% & 6.1e-4 & 2.1e-2 & 0.39\% & 1.65\% & 1.6e-2 & 2.7e-2 \\ \hline
\hline
\multirow{5}{*}{5} & 1 & 0.01\% & 1.85\% & 2.0e-1 & 6.9e-2 & 0.37\% & 3.42\% & 1.2e-1 & 1.9e-1 \\ \cline{2-10} 
 & 5 & 0.01\% & \textbf{1.83\%} & 5.9e-2 & 9.3e-2 & 0.0\% & 2.31\% & 1.0e-2 & 1.1e-1 \\ \cline{2-10} 
 & 10 & 0.0\% & 2.31\% & 1.3e-2 & 1.1e-1 & 0.0\% & 2.10\% & 1.1e-3 & 1.0e-1 \\ \cline{2-10} 
 & 20 & 0.0\% & 2.11\% & 6.0e-4 & 1.0e-1 & 0.0\% & 2.06\% & 1.0e-4 & 1.0e-1 \\ \cline{2-10} 
 & 40 & 0.0\% & 2.09\% & 0 & 1.0e-1 & 0.0\% & 2.07\% & 0 & 1.0e-1 \\ \hline
\end{tabular}
}
\end{adjustbox}
\end{table}

\end{appendices}

\end{document}